\documentclass{article}

\PassOptionsToPackage{numbers, compress}{natbib}

    \usepackage[preprint]{neurips_2019}

\usepackage[utf8]{inputenc} 
\usepackage[T1]{fontenc}    
\usepackage{hyperref}       
\usepackage{url}            
\usepackage{booktabs}       
\usepackage{amsfonts}       
\usepackage{nicefrac}       
\usepackage{microtype}      
\usepackage{graphicx}
\usepackage{subfig}
\usepackage{booktabs} 
\usepackage{ amssymb }
\usepackage{amsthm}
\usepackage{amsmath}
\usepackage{arydshln}
\usepackage{xcolor}
\usepackage[utf8]{inputenc}
\usepackage{url}
\usepackage[demo]{rotating}
\usepackage{amssymb}
\usepackage{bbm}
\usepackage{rotating}
\usepackage{esvect}

\newtheorem{proposition}{Proposition}
\newtheorem{exampl}{Example}

\newtheorem{defn}{Definition}
\newenvironment{definition}{\begin{defn}\em}{\end{defn}}
\newtheorem{theorem}{Theorem}

\newcommand{\vctr}[1]{\ensuremath{\pmb{#1}}}
\newcommand{\mtrx}[1]{\ensuremath{\pmb{#1}}}
\newcommand{\graph}[1]{\ensuremath{\mathcal{#1}}}
\newcommand{\vertices}[1]{\ensuremath{\mathcal{#1}}}

\newcommand{\vertex}[1]{\ensuremath{\mathsf{#1}}}

\newcommand{\timestamp}[1]{\ensuremath{\mathsf{#1}}}
\newcommand{\relations}[1]{\ensuremath{\mathcal{#1}}}
\newcommand{\timestamps}[1]{\ensuremath{\mathcal{#1}}}
\newcommand{\relation}[1]{\ensuremath{\mathsf{#1}}}

\newcommand{\function}[1]{\ensuremath{\mathtt{#1}}}

\newcommand{\Th}[1]{\ensuremath{#1^{\text{th}}}}
\newcommand{\transpose}[1]{\ensuremath{#1^{\textsf{T}}}}

\newcommand{\world}[1]{\ensuremath{\mathcal{#1}}}
\newcommand{\sumElemProd}[1]{\ensuremath{\langle#1\rangle}}

\newcommand{\cev}[1]{\reflectbox{\ensuremath{\vec{\reflectbox{\ensuremath{#1}}}}}}

\title{Diachronic Embedding for Temporal Knowledge Graph Completion}

\author{%
  Rishab Goel\thanks{Equal contribution.}~, Seyed Mehran Kazemi$^*$, Marcus Brubaker, Pascal Poupart \\
  Borealis AI \\
  \texttt{\{rishab.goel,mehran.kazemi,marcus.brubaker,pascal.poupart\}@borealisai.com} \\
}

\begin{document}

\maketitle

\begin{abstract}
Knowledge graphs (KGs) typically contain temporal facts indicating relationships among entities at different times. Due to their incompleteness, several approaches have been proposed to infer new facts for a KG based on the existing ones--a problem known as \emph{KG completion}. KG embedding approaches have proved effective for KG completion, however, they have been developed mostly for static KGs. 
Developing temporal KG embedding models is an increasingly important problem. In this paper, we build novel models for temporal KG completion through equipping static models with a diachronic entity embedding function which provides the characteristics of entities at \emph{any} point in time. 
This is in contrast to the existing temporal KG embedding approaches where only static entity features are provided.
The proposed embedding function is model-agnostic and can be potentially combined with any static model. We prove that combining it with SimplE, a recent model for static KG embedding, results in a fully expressive model for temporal KG completion. Our experiments indicate the superiority of our proposal compared to existing baselines.
\end{abstract}

\section{Introduction}
Knowledge graphs (KGs) are directed graphs where nodes represent entities and (labeled) edges represent the types of relationships among entities. Each edge in a KG corresponds to a fact and can be represented as a tuple such as 
$(\vertex{Mary}, \relation{Liked}, \vertex{God~Father})$ where \vertex{Mary} and \vertex{God~Father} are called the head and tail entities respectively and \relation{Liked} is a relation. 
An important problem, known as KG completion, is to infer new facts from a KG based on the existing ones. This problem has been extensively studied for static KGs (see \cite{nickel2016review,wang2017knowledge,nguyen2017overview} for a summary). 
KG embedding approaches have offered state-of-the-art results for KG completion on several benchmarks. These approaches map each entity and each relation type to a hidden representation and compute a score for each tuple by applying a score function to these representations.
Different approaches differ in how they map the entities and relation types to hidden representations and in their score functions.

To capture the temporal aspect of the facts, KG edges are typically associated with a timestamp or time interval; e.g., $(\vertex{Mary}, \relation{Liked}, \vertex{God~Father}, 1995)$. However, KG embedding approaches have been mostly designed for static KGs ignoring the temporal aspect.
Recent work has shown a substantial boost in performance by extending these approaches to utilize time \cite{jiang2016towards,dasgupta2018hyte,ma2018embedding,garcia2018learning}. The proposed extensions are mainly through computing a hidden representation for each timestamp and extending the score functions to utilize timestamp representations as well as entity and relation representations.

In this paper, we develop models for temporal KG completion (TKGC) based on an intuitive assumption: to provide a score for, e.g., $(\vertex{Mary}, \relation{Liked}, \vertex{God~Father}, \timestamp{1995})$, one needs to know \vertex{Mary}'s and \vertex{God~Father}'s features on $\timestamp{1995}$; providing a score based on their current features may be misleading. That is because $\vertex{Mary}$'s personality and the sentiment towards \vertex{God~Father} may have been quite different on \timestamp{1995} as compared to now. Consequently, learning a static representation for each entity -- as is done by existing approaches -- may be sub-optimal as such a representation only captures the entity features at the current time, or an aggregation of entity features during time.

To provide entity features at any given time, we define entity embedding as a function which takes an entity and a timestamp as input and provides a hidden representation for the entity at that time. Inspired by diachronic word embeddings, we call our proposed embedding \emph{diachronic embedding (DE)}. DE is model-agnostic: any static KG embedding model can be potentially extended to TKGC by leveraging DE. We prove that combining DE with SimplE \cite{kazemi2018simple} results in a fully expressive model for TKGC. To the best of our knowledge, this is the first TKGC model with a proof of fully expressiveness. We show the merit of our model on
subsets of ICEWS \cite{boschee2015icews} and 
GDELT \cite{leetaru2013gdelt} datasets. 
 
\section{Background and Notation} \label{sec:background}
\textbf{Notation:} Lower-case letters denote scalars, bold lower-case letters  denote vectors, and bold upper-case letters denote matrices. $\vctr{z}[n]$ represents the $\Th{n}$ element of a vector \vctr{z},  $||\vctr{z}||$ represents its norm, and $\transpose{\vctr{z}}$ represents its transpose. For two vectors $\vctr{z_1}\in\mathbb{R}^{d_1}$ and $\vctr{z_2}\in\mathbb{R}^{d_2}$,  $[\vctr{z}_1;\vctr{z}_2]\in\mathbb{R}^{d_1+d_2}$ represents the concatenation of the two vectors. $\vctr{z}_1 \otimes \vctr{z}_2$ represents a vector $\vctr{z}\in\mathbb{R}^{d_1 d_2}$ such that $\vctr{z}[(n-1) * d_2 + m]=\vctr{z}_1[n] * \vctr{z}_2[m]$ (i.e. the flattened vector of the tensor/outer product of the two vectors). For $k$ vectors $\vctr{z}_1, \dots, \vctr{z}_k$ of the same length $d$, $\sumElemProd{\vctr{z}_1, \dots, \vctr{z}_k}=\sum_{n=1}^d (\vctr{z}_1[n] * \dots * \vctr{z}_k[n])$ represents the sum of the element-wise product of the elements of the $k$ vectors.

\textbf{Temporal Knowledge Graph (Completion):} Let $\vertices{V}$ be a finite set of entities, $\relations{R}$ be a finite set of relation types, and $\timestamps{T}$ be a finite set of timestamps.
Let $\world{W}\subset \vertices{V}\times\relations{R}\times\vertices{V}\times\timestamps{T}$ represent the set of all temporal tuples $(\vertex{v}, \relation{r}, \vertex{u}, \timestamp{t})$ that are facts (i.e. true in a world), where $\vertex{v},\vertex{u}\in\vertices{V}$, $\relation{r}\in\relations{R}$, and $\timestamp{t}\in\timestamps{T}$. Let $\world{W}^c$ be the complement of $\world{W}$. A temporal knowledge graph (KG) $\graph{G}$ is a subset of $\world{W}$ (i.e. $\graph{G}\subset \world{W}$). Temporal KG completion (TKGC) is the problem of inferring \world{W} from \graph{G}. 

\textbf{Relation Properties:} A relation $\relation{r}$ is \emph{symmetric} if $(\vertex{v}, \relation{r}, \vertex{u}, \timestamp{t}) \in \world{W} \iff (\vertex{u}, \relation{r}, \vertex{v}, \timestamp{t}) \in \world{W}$ and \emph{anti-symmetric} if $(\vertex{v}, \relation{r}, \vertex{u}, \timestamp{t}) \in \world{W} \iff (\vertex{u}, \relation{r}, \vertex{v}, \timestamp{t}) \in \world{W}^c$. A relation $\relation{r}_i$ is the inverse of another relation $\relation{r}_j$ if $(\vertex{v}, \relation{r}_i, \vertex{u}, \timestamp{t}) \in \world{W} \iff (\vertex{u}, \relation{r}_j, \vertex{v}, \timestamp{t}) \in \world{W}$. $\relation{r}_i$ entails $\relation{r}_j$ if $(\vertex{v}, \relation{r}_i, \vertex{u}, \timestamp{t}) \in \world{W} \Rightarrow (\vertex{v}, \relation{r}_j, \vertex{u}, \timestamp{t}) \in \world{W}$.

\textbf{KG Embedding:} Formally, we define an entity embedding as follows.
\begin{definition} \label{embedding-dfnt}
An \emph{entity embedding}, $\function{EEMB}: \vertices{V} \rightarrow \psi$, is a function which maps every entity $\vertex{v}\in\vertices{V}$ to a hidden representation in $\psi$ where $\psi$ is the class of non-empty tuples of vectors and/or matrices.
\end{definition}
A \emph{relation embedding} ($\function{REMB}: \relations{R} \rightarrow \psi$) is defined similarly. We refer to the hidden representation of an entity (relation) as the embedding of the entity (relation). A KG embedding model defines two things: 1- the $\function{EEMB}$ and $\function{REMB}$ functions, 2- a score function which takes $\function{EEMB}$ and $\function{REMB}$ as input and provides a score for a given tuple. The parameters of hidden representations are learned from data.

\section{Existing Approaches} 
In this section, we describe the existing approaches for static and temporal KG completion that will be used in the rest of the paper. For further detail on temporal KG completion approaches, we refer the reader to a recent survey \cite{kazemi2019relational}. We represent the score for a tuple by $\phi(.)$.

\textbf{TransE (static) \cite{bordes2013translating}:} In TransE, $\function{EEMB}(\vertex{v})=(\vctr{z}_\vertex{v})$ for every $\vertex{v}\in\vertices{V}$ where $\vctr{z}_\vertex{v}\in\mathbb{R}^d$, $\function{REMB}(\vertex{r})=(\vctr{z}_\vertex{r})$ for every $\vertex{r}\in\vertices{R}$ where $\vctr{z}_\vertex{r}\in\mathbb{R}^d$, and $\phi(\vertex{v}, \relation{r}, \vertex{u})=-|| \vctr{z}_\vertex{v} + \vctr{z}_\relation{r} - \vctr{z}_\vertex{u}||$.

\textbf{DistMult (static) \cite{yang2015embedding}:} Same $\function{EEMB}$ and $\function{REMB}$ as TransE but defining $\phi(\vertex{v}, \relation{r}, \vertex{u})= \sumElemProd{\vctr{z}_\vertex{v}, \vctr{z}_\relation{r}, \vctr{z}_\vertex{u}}$.

\textbf{Tucker (static) \cite{tucker1966some,balavzevic2019tucker}:} Same $\function{EEMB}$ and $\function{REMB}$ as TransE but defining $\phi(\vertex{v}, \relation{r}, \vertex{u})=\sumElemProd{\vctr{w}, \vctr{z}_\vertex{v} \otimes \vctr{z}_\relation{r} \otimes \vctr{z}_\vertex{u}}$ where $\vctr{w}\in\mathbb{R}^{d^3}$ is a weight vector shared for all tuples.

\textbf{RESCAL (static) \cite{nickel2011three}:} Same $\function{EEMB}$ as TransE but defining  $\function{REMB}(\relation{r})=(\mtrx{Z}_\relation{r})$ for every $\relation{r}\in\relations{R}$ where $\mtrx{Z}_\relation{r}\in\mathbb{R}^{d\times d}$, and defining $\phi(\vertex{v}, \relation{r}, \vertex{u})=\transpose{\vctr{z}}_\vertex{v} \vctr{Z}_\relation{r} \vctr{z}_\vertex{u}$.

\textbf{Canonical Polyadic (CP) (static) \cite{hitchcock1927expression}:} Same $\function{REMB}$ as TransE but defining  $\function{EEMB}(\vertex{v})=(\vec{\vctr{z}}_\vertex{v}, \cev{\vctr{z}}_\vertex{v})$ for every $\vertex{v}\in\vertices{V}$ where $\vec{\vctr{z}}_\vertex{v}, \cev{\vctr{z}}_\vertex{v}\in\mathbb{R}^d$. $\vec{\vctr{z}}_\vertex{v}$ is used when $\vertex{v}$ is the head and $\cev{\vctr{z}}_\vertex{v}$ is used when $\vertex{v}$ is the tail. In CP,  $\phi(\vertex{v}, \relation{r}, \vertex{u})=\sumElemProd{\vec{\vctr{z}}_\vertex{v}, \vctr{z}_\vertex{r}, \cev{\vctr{z}}_\vertex{u}}$. DistMult is a special case of CP where $\vec{\vctr{z}}_\vertex{v}=\cev{\vctr{z}}_\vertex{v}$ for every $\vertex{v}\in\vertices{V}$.

\textbf{SimplE (static) \cite{kazemi2018simple}:} Noticing an information flow issue between the two vectors $\vec{\vctr{z}}_\vertex{v}$ and $\cev{\vctr{z}}_\vertex{v}$ of an entity $\vertex{v}$ in CP, \citet{kazemi2018simple} take advantage of the inverse of the relations to address this issue. They define $\function{REMB}(\vertex{r})=(\vec{\vctr{z}}_\vertex{r}, \cev{\vctr{z}}_\vertex{r})$ for every $\vertex{r}\in\vertices{R}$, where $\vec{\vctr{z}}_\vertex{r}$ is used as in CP and $\cev{\vctr{z}}_\vertex{r}\in\mathbb{R}^d$ is considered the embedding of $\relation{r}^{-1}$, the inverse of $\relation{r}$. In SimplE, $\phi(\vertex{v}, \relation{r}, \vertex{u})$ is defined as the average of two CP scores: 1- $\sumElemProd{\vec{\vctr{z}}_\vertex{v}, \vec{\vctr{z}}_\vertex{r}, \cev{\vctr{z}}_\vertex{u}}$ corresponding to the score for $(\vertex{v}, \relation{r}, \vertex{u})$ and 2- $\sumElemProd{\vec{\vctr{z}}_\vertex{u}, \cev{\vctr{z}}_\vertex{r}, \cev{\vctr{z}}_\vertex{v}}$ corresponding to the score for $(\vertex{u}, \relation{r}^{-1}, \vertex{v})$. A similar extension of CP has been proposed in \cite{lacroix2018canonical}.

\textbf{TTransE (temporal) \cite{jiang2016towards}:} An extension of TransE by adding one more embedding function mapping timestamps to hidden representations: $\function{TEMB}(\timestamp{t})=(\vctr{z}_\timestamp{t})$ for every $\timestamp{t}\in\timestamps{T}$ where $\vctr{z}_\timestamp{t}\in\mathbb{R}^d$. In TTransE, $\phi(\vertex{v}, \relation{r}, \vertex{u},\timestamp{t})=-|| \vctr{z}_\vertex{v} + \vctr{z}_\relation{r} + \vctr{z}_\timestamp{t} - \vctr{z}_\vertex{u} ||$.

\textbf{HyTE (temporal) \cite{dasgupta2018hyte}:} Same $\function{EEMB}$, $\function{REMB}$ and $\function{TEMB}$ as TTransE but defining $\phi(\vertex{v}, \relation{r}, \vertex{u},\timestamp{t})=-|| \overline{\vctr{z}}_\vertex{v} + \overline{\vctr{z}}_\relation{r} - \overline{\vctr{z}}_\vertex{u} ||$ where $\overline{\vctr{z}}_\vertex{x}=\vctr{z}_\vertex{x} - \transpose{\vctr{z}}_\timestamp{t}\vctr{z}_\vertex{x}\vctr{z}_\timestamp{t}$ for $\vertex{x}\in\{\vertex{v}, \relation{r}, \vertex{u}\}$. Intuitively, HyTE first projects the head, relation, and tail embeddings to the space of the timestamp and then applies the TransE function on the projected embeddings. 

\textbf{ConT (temporal) \cite{ma2018embedding}:} \citet{ma2018embedding} extend several static KG embedding models to TKGC. Their best performing model, ConT, is an extension of Tucker defining $\function{TEMB}(\timestamp{t})=(\vctr{z}_\timestamp{t})$ for every $\timestamp{t}\in\timestamps{T}$ where $\vctr{z}_\timestamp{t}\in\mathbb{R}^{d^3}$ and changing the score function to $\phi(\vertex{v}, \relation{r}, \vertex{u},\timestamp{t})=\sumElemProd{\vctr{z}_\timestamp{t}, \vctr{z}_\vertex{v} \otimes \vctr{z}_\relation{r} \otimes \vctr{z}_\vertex{u}}$. Intuitively, ConT replaces the shared vector $\vctr{w}$ in Tucker with timestamp embeddings $\vctr{z}_\timestamp{t}$.

\textbf{TA-DistMult (temporal) \cite{garcia2018learning}:} An extension of DistMult where each character $c$ in the timestamps is mapped to a vector ($\function{CEMB}(c)=\vctr{z}_c$) where $\vctr{z}_c\in\mathbb{R}^d$. Then, for a tuple $(\vertex{v}, \relation{r}, \vertex{u}, \timestamp{t})$, a temporal relation is created by considering $\relation{r}$ and the characters in $\timestamp{t}$ as a sequence and an embedding $\vctr{z}_{\relation{r},\timestamp{t}}$ is computed for this temporal relation by feeding the embedding vectors for each element in the sequence to an LSTM and taking its final output. Finally, the score function of DistMult is employed: $\phi(\vertex{v}, \relation{r}, \vertex{u}, \timestamp{t})=\sumElemProd{\vctr{z}_\vertex{v},\vctr{z}_{\relation{r},\timestamp{t}},\vctr{z}_\vertex{u}}$ (TransE was employed as well but DistMult performed better).

\section{Diachronic Embedding} \label{sec:de-emb}
According to Definition~\ref{embedding-dfnt},
an entity embedding function takes an entity as input and provides a hidden representation as output. We propose an alternative entity embedding function which, besides entity, takes time as input as well. Inspired by diachronic word embeddings, we call such an embedding function a \emph{diachronic entity embedding}. Below is a formal definition of a diachronic entity embedding.

\begin{definition} \label{de-embedding-dfnt}
A \emph{diachronic entity embedding}, $\function{DEEMB}: (\vertices{V}, \timestamps{T}) \rightarrow \psi$, is a function which maps every pair $(\vertex{v}, \timestamp{t})$, where $\vertex{v}\in\vertices{V}$ and $\timestamp{t}\in\timestamps{T}$, to a hidden representation in $\psi$ where $\psi$ is the class of non-empty tuples of vectors and/or matrices.
\end{definition}
 
One may take their favorite static KG embedding score function and make it temporal by replacing their entity embeddings with diachronic entity embeddings. The choice of the $\function{DEEMB}$ function can be different for various temporal KGs depending on their properties. Here, we propose a \function{DEEMB} function which performs well on our benchmarks. We give the definition for models where the output of the $\function{DEEMB}$ function is a tuple of vectors but it can be generalized to other cases as well. Let $\vctr{z}^\timestamp{t}_\vertex{v}\in\mathbb{R}^d$ be a vector in $\function{DEEMB}(\vertex{v}, \timestamp{t})$ (i.e. $\function{DEEMB}(\vertex{v}, \timestamp{t})=(\dots,\vctr{z}^\timestamp{t}_\vertex{v},\dots)$). We define $\vctr{z}^\timestamp{t}_\vertex{v}$ as follows:
\begin{equation}
    \label{eq:demb}
  \vctr{z}^\timestamp{t}_\vertex{v}[n]=\begin{cases}
    \vctr{a}_\vertex{v}[n] \sigma(\vctr{w}_\vertex{v}[n] \timestamp{t} + \vctr{b}_\vertex{v}[n]), & \text{if~~$1 \leq n\leq \gamma d$}. \\
    \vctr{a}_\vertex{v}[n], & \text{if~~$\gamma d < n \leq d$}.
  \end{cases}
\end{equation}
where $\vctr{a}_\vertex{v}\in\mathbb{R}^{d}$ and $\vctr{w}_\vertex{v},\vctr{b}_\vertex{v}\in\mathbb{R}^{\gamma d}$ are (entity-specific) vectors with learnable parameters and $\sigma$ is an activation function. Intuitively, entities may have some features that change over time and some features that remain fixed. 
The first $\gamma d$ elements of the vector in Equation~\eqref{eq:demb} capture temporal features 
and the other $(1-\gamma)d$ elements capture static features. $0\leq \gamma \leq 1$ is a hyper-parameter controlling the percentage of temporal features.
While in Equation~\eqref{eq:demb} static features can be potentially obtained from the temporal ones if the optimizer sets some elements of $\vctr{w}_\vertex{v}$ to zero, explicitly modeling static features helps reduce the number of learnable parameters and avoid overfitting to temporal signals (see Section~\ref{sec:ablation}).

Intuitively, by learning $\vctr{w}_\vertex{v}$s and $\vctr{b}_\vertex{v}$s, the model learns how to turn entity features on and off at different points in time so accurate temporal predictions can be made about them at any time. $\vctr{a}_\vertex{v}$s control the importance of the features. We mainly use \emph{sine} as the activation function for Equation~\eqref{eq:demb} because one sine function can model several on and off states. Our experiments explore other activation functions as well and provide more intuition.

\textbf{Model-Agnosticism:} The proposals in existing temporal KG embedding models can only extend one (or a few) static models to temporal KGs. As an example, it is not trivial how RESCAL can be extended to temporal KGs using the proposal in \cite{garcia2018learning} (except for the naive approach of expecting the LSTM to output large $\mtrx{Z}_\relation{r}$ matrices) or in \cite{jiang2016towards,dasgupta2018hyte}. Same goes for models other than RESCAL where the relation embeddings contain matrices (see, e.g., \cite{StransE,socher2013reasoning,lin2015learning}). Using our proposal, one may construct temporal versions of TransE, DistMult, SimplE, Tucker, RESCAL, or other models by replacing their $\function{EEMB}$ function with $\function{DEEMB}$ in Equation~\ref{eq:demb}. We refer to the resulting models as \emph{DE-TransE}, \emph{DE-DistMult}, \emph{DE-SimplE} and so forth, where \emph{DE} is short for \textbf{D}iachronic \textbf{E}mbedding.

\textbf{Learning:} The facts in a KG $\graph{G}$ are split into $train$, $validation$, and $test$ sets. Model parameters are learned using stochastic gradient descent with mini-batches. Let $B\subset train$ be a mini-batch. For each fact $f=(\vertex{v}, \relation{r}, \vertex{u}, \timestamp{t})\in B$, we generate two queries: 1- $(\vertex{v}, \relation{r}, ?, \timestamp{t})$ and 2- $(?, \relation{r}, \vertex{u}, \timestamp{t})$. For the first query, we generate a candidate answer set $\mathsf{C}_{f,\vertex{v}}$ which contains $\vertex{v}$ and $n$ (hereafter referred to as negative ratio) other entities selected randomly from $\vertices{V}$. For the second query, we generate a similar candidate answer set $\mathsf{C}_{f,\vertex{u}}$. Then we minimize the cross entropy loss which has been used and shown good results for both static and temporal KG completion (see, e.g., \cite{kadlec2017knowledge,garcia2018learning}):
\begin{align*}
    \mathcal{L}=-\big(\sum_{f=(\vertex{v}, \relation{r}, \vertex{u}, \timestamp{t})\in B} \frac{\exp(\phi(\vertex{v}, \relation{r}, \vertex{u}, \timestamp{t}))}{\sum_{\vertex{u}'\in \mathsf{C}_{f,\vertex{u}}} \exp(\phi(\vertex{v}, \relation{r}, \vertex{u}', \timestamp{t}))} + \frac{\exp(\phi(\vertex{v}, \relation{r}, \vertex{u}, \timestamp{t}))}{\sum_{\vertex{v}'\in \mathsf{C}_{f,\vertex{v}}} \exp(\phi(\vertex{v}', \relation{r}, \vertex{u}, \timestamp{t}))}\big)
\end{align*}

\subsection{Expressivity}
Expressivity is an important property and has been the subject of study in several recent works on static (knowledge) graphs \cite{buchman2016negation,trouillon2017knowledge,kazemi2018simple,xu2019powerful,balavzevic2019tucker,fatemi2019improved}. If a model is not expressive enough, it is doomed to underfitting for some applications. A desired property of a model is fully expressiveness:
 
\begin{definition} \label{full-exp-dfnt}
A model with parameters $\theta$ is \emph{fully expressive} if given any world with true tuples $\world{W}$ and false tuples $\world{W}^c$, there exists an assignment for $\theta$ that correctly classifies the tuples in $\world{W}$ and $\world{W}^c$.
\end{definition}

For static KG completion, several models have been proved to be fully expressive. For TKGC, however, a proof of fully expressiveness does not yet exist for the proposed models. The following theorem establishes the fully expressiveness of DE-SimplE. The proof can be found in Appendix A.

\begin{theorem}[\textbf{Expressivity}]
DE-SimplE is fully expressive for temporal knowledge graph completion.
\end{theorem}

\subsection{Domain Knowledge}
For several static KG embedding models, it has been shown how certain types of domain knowledge (if exists) can be incorporated into the embeddings through parameter sharing (aka tying) and how it helps improve model performance (see, e.g., \cite{kazemi2018simple,sun2019rotate,minervini2017regularizing,fatemi2019improved}). 
Incorporating domain knowledge for these static models can be ported to their temporal version when they are extended to temporal KGs through our diachronic embeddings. As a proof of concept, we show how incorporating domain knowledge into SimplE can be ported to DE-SimplE. We chose SimplE for our proof of concept as several types of domain knowledge can be incorporated into it.  

Consider $\relation{r}_i\in\relations{R}$ with $\function{REMB}(\relation{r}_i)=(\vec{\vctr{z}}_{\relation{r}_i}, \cev{\vctr{z}}_{\relation{r}_i})$ (according to SimplE).
If $\relation{r}_i$ is known to be symmetric or anti-symmetric, this knowledge can be incorporated into the embeddings by tying $\vec{\vctr{z}}_{\relation{r}_i}$ to $\cev{\vctr{z}}_{\relation{r}_i}$ or negation of $\cev{\vctr{z}}_{\relation{r}_i}$ respectively \cite{kazemi2018simple}. 
If $\relation{r}_i$ is known to be the inverse of $\relation{r}_j$, this knowledge can be incorporated into the embeddings by tying $\vec{\vctr{z}}_{\relation{r}_i}$ to $\cev{\vctr{z}}_{\relation{r}_j}$ and $\vec{\vctr{z}}_{\relation{r}_j}$ to $\cev{\vctr{z}}_{\relation{r}_i}$ \cite{kazemi2018simple}.

\begin{proposition} \label{prop:sym-asym-inv}
Symmetry, anti-symmetry, and inversion can be incorporated into DE-SimplE in the same way as SimplE.
\end{proposition}

If $\relation{r}_i$ is known to entail $\relation{r}_j$, \citet{fatemi2019improved} prove that if entity embeddings are constrained to be non-negative, then this knowledge can be incorporated by tying $\vec{\vctr{z}}_{\relation{r}_j}$ to $\vec{\vctr{z}}_{\relation{r}_i}+\vec{\vctr{\delta}}_{\relation{r}_j}$ and  $\cev{\vctr{z}}_{\relation{r}_j}$ to $\cev{\vctr{z}}_{\relation{r}_i}+\cev{\vctr{\delta}}_{\relation{r}_j}$ where $\vec{\vctr{\delta}}_{\relation{r}_j}$ and $\cev{\vctr{\delta}}_{\relation{r}_j}$ are vectors with non-negative elements. We give a similar result for DE-SimplE.

\begin{proposition} \label{prop:subsum}
By constraining $\vctr{a}_\vertex{v}$s in Equation~\eqref{eq:demb} to be non-negative for all $\vertex{v}\in\vertices{V}$ and $\sigma$ to be an activation function with a non-negative range (such as ReLU, sigmoid, or squared exponential), entailment can be incorporated into DE-SimplE in the same way as SimplE.
\end{proposition}

Compared to the result in \citet{fatemi2019improved}, the only added constraint for DE-SimplE is that the activation function in Equation~\eqref{eq:demb} is also constrained to have a non-negative range. Proofs for Propositions~\ref{prop:sym-asym-inv} and \ref{prop:subsum} can be found in Appendix A.

\begin{table}[b!]
\footnotesize
\caption{Statistics on ICEWS14, ICEWS05-15, and GDELT.}
\label{tab:dataset-stats}
\begin{center}
\begin{tabular}{c||c|c|c|c|c|c|c}
Dataset & $|\vertices{V}|$ & $|\relations{R}|$ & |\timestamps{T}| & $|train|$ & $|validation|$ & $|test|$ & $|\graph{G}|$ \\ \hline
ICEWS14 & 7,128 & 230 & 365 & 72,826 & 8,941 & 8,963 & 90,730 \\
ICEWS05-15 & 10,488 & 251 & 4017 & 386,962 & 46,275 & 46,092 & 479,329\\
GDELT & 500 & 20 & 366 & 2,735,685 & 341,961 & 341,961 & 3,419,607
\end{tabular}
\end{center}
\end{table}

\section{Experiments \& Results}
\textbf{Datasets:} Our datasets are subsets of two temporal KGs that have become standard benchmarks for TKGC: ICEWS \cite{boschee2015icews} and GDELT \cite{leetaru2013gdelt}. For ICEWS, we use the two subsets generated by \citet{garcia2018learning}: 1- \emph{ICEWS14} corresponding to the facts in 2014 and 2- \emph{ICEWS05-15} corresponding to the facts between 2005 to 2015. For GDELT, we use the subset extracted by \citet{trivedi2017know} corresponding to the facts from April 1, 2015 to March 31, 2016. We changed the train/validation/test sets following a similar procedure as in \cite{bordes2013translating} to make the problem into a TKGC rather than an extrapolation problem. Table~\ref{tab:dataset-stats} provides a summary of the dataset statistics.

\textbf{Baselines:} Our baselines include both static and temporal KG embedding models. From the static KG embedding models, we use TransE and DistMult and SimplE where the timing information are ignored. From the temporal KG embedding models, we use the ones introduced in Section~\ref{sec:background}.

\textbf{Metrics:} For each fact $f=(\vertex{v}, \relation{r}, \vertex{u}, \timestamp{t})\in test$, we create two queries: 1- $(\vertex{v}, \relation{r}, ?, \timestamp{t})$ and 2- $(?, \relation{r}, \vertex{u}, \timestamp{t})$. For the first query, the model ranks all entities in $\vertex{u}\cup\overline{\mathsf{C}}_{f,\vertex{u}}$ where $\overline{\mathsf{C}}_{f,\vertex{u}}=\{\vertex{u}':\vertex{u}'\in\vertices{V}, (\vertex{v}, \relation{r}, \vertex{u}', \timestamp{t})\not\in \graph{G}\}$.  This corresponds to the filtered setting commonly used in the literature \cite{bordes2013translating}. We follow a similar approach for the second query. Let $k_{f,\vertex{u}}$ and $k_{f, \vertex{v}}$ represent the ranking for $\vertex{u}$ and $\vertex{v}$ for the two queries respectively. We report \emph{mean reciprocal rank (MRR)} defined as $\frac{1}{2*|test|}\sum_{f=(\vertex{v},\relation{r},\vertex{u},\timestamp{t})\in test}(\frac{1}{k_{f,\vertex{u}}}+\frac{1}{k_{f, \vertex{v}}})$. Compared to its counterpart \emph{mean rank} which is largely influenced by a single bad prediction, MRR is more stable \cite{nickel2016review}. We also report Hit@1, Hit@3 and Hit@10 measures where Hit@k is defined as $\frac{1}{2*|test|}\sum_{f=(\vertex{v},\relation{r},\vertex{u},\timestamp{t})\in test} (\mathbbm{1}_{k_{f,\vertex{u}} \leq k} + \mathbbm{1}_{k_{f,\vertex{v}} \leq k})$, where $\mathbbm{1}_{cond}$ is $1$ if $cond$ holds and $0$ otherwise.

\textbf{Implementation\footnote{Code and datasets are available at \url{https://github.com/BorealisAI/DE-SimplE}}:} We implemented our model and the baselines in PyTorch \cite{paszke2017automatic}. We ran our experiments on a node with four GPUs. 
For the two ICEWS datasets, we report the results for some of the baselines from \cite{garcia2018learning}. For the other experiments on these datasets, for the fairness of results, we follow a similar experimental setup as in \cite{garcia2018learning} by using the ADAM optimizer \cite{kingma2014adam} and setting learning rate $=0.001$, batch size $=512$, negative ratio $=500$, embedding size $=100$, and validating every $20$ epochs selecting the model giving the best validation MRR. 
Following the best results obtained in \cite{ma2018embedding} (and considering the memory restrictions), for ConT we set embedding size $=40$, batch size $=32$ on ICEWS14 and GDELT and $16$ on ICEWS05-15. We validated dropout values from $\{0.0, 0.2, 0.4\}$. We tuned $\gamma$ for our model from the values $\{16, 32, 64\}$. For GDELT, we used a similar setting but with a negative ratio $=5$ due to the large size of the dataset. Unless stated otherwise, we use $sine$ as the activation function for Equation~\eqref{eq:demb}. Since the timestamps in our datasets are dates rather than single numbers, we apply the temporal part of Equation~\eqref{eq:demb} to year, month, and day separately (with different parameters) thus obtaining three temporal vectors. Then we take an element-wise sum of the resulting vectors obtaining a single temporal vector. Intuitively, this can be viewed as converting a date into a timestamp in the embedded space. 

\subsection{Comparative Study}
We compare the baselines with three variants of our model: 1- DE-TransE, 2- DE-DistMult, and 3- DE-SimplE. The obtained results in Table~\ref{tab:results} indicate that the large number of parameters per timestamp makes ConT perform poorly on ICEWS14 and ICEWS05-15. On GDELT, it shows a somewhat better performance as GDELT has many training facts in each timestamp. Besides affecting the predictive performance, the large number of parameters makes training ConT extremely slow. 
According to the results, the temporal versions of different models outperform the static counterparts in most cases, thus providing evidence for the merit of capturing temporal information.

DE-TransE outperforms the other TransE-based baselines (TTransE and HyTE) on ICEWS14 and GDELT and gives on-par results with HyTE on ICEWS05-15. This result shows the superiority of our diachronic embeddings compared to TTransE and HyTE. DE-DistMult outperforms TA-DistMult, the only DistMult-based baseline,  showing the superiority of our diachronic embedding compared to TA-DistMult. Moreover, DE-DistMult outperforms all TransE-based baselines.  Finally, just as SimplE beats TransE and DistMult due to its higher expressivity, our results show that DE-SimplE beats DE-TransE, DE-DistMult, and the other baselines due to its higher expressivity.

Previously, each of the existing models was tested on different subsets of ICEWS and GDELT and a comprehensive comparison of them did not exist. As a side contribution, Table~\ref{tab:results} provides a comparison of these approaches on the same benchmarks and under the same experimental setting. The results reported in Table~\ref{tab:results} may be directly used for comparison in future works.

\begin{table*}[t]
\footnotesize
\setlength{\tabcolsep}{2pt}
\caption{Results on ICEWS14, ICEWS05-15, and GDELT. Best results are in bold.}
\label{tab:results}
\begin{center}
\begin{tabular}{c|cccc|cccc|cccc}
\toprule
& \multicolumn{4}{c}{ICEWS14} & \multicolumn{4}{c}{ICEWS05-15} &  \multicolumn{4}{c}{GDELT}  \\
\cmidrule(lr){2-5} \cmidrule(lr){6-9} \cmidrule(lr){10-13}
Model & MRR & Hit@1 & Hit@3 & Hit@10 & MRR & Hit@1 & Hit@3 & Hit@10 & MRR & Hit@1 & Hit@3 & Hit@10 \\ \hline
TransE & 0.280 & 9.4 & - & 63.7 & 0.294 & 9.0 & - & 66.3 & 0.113 & 0.0 & 15.8 & 31.2 \\
DistMult & 0.439 & 32.3 & - & 67.2 & 0.456 & 33.7 & - & 69.1 & 0.196 & 11.7 & 20.8 & 34.8 \\
SimplE & 0.458 & 34.1 & 51.6 & 68.7 & 0.478 & 35.9 & 53.9 & 70.8 & 0.206 & 12.4 & 22.0 & 36.6 \\
ConT & 0.185 & 11.7 & 20.5 & 31.5 & 0.163 & 10.5 & 18.9 & 27.2 & 0.144 & 8.0 & 15.6 & 26.5 \\
TTransE & 0.255 & 7.4 & - & 60.1 & 0.271 & 8.4 & - & 61.6 & 0.115 & 0.0 & 16.0 & 31.8 \\
HyTE & 0.297 & 10.8 & 41.6 & 65.5 & 0.316 & 11.6 & 44.5 & 68.1 & 0.118 & 0.0 & 16.5 & 32.6 \\
TA-DistMult & 0.477 & 36.3 & - & 68.6 & 0.474 & 34.6 & - & 72.8 & 0.206 & 12.4 & 21.9 & 36.5 \\ \hline
DE-TransE & 0.326 & 12.4 & 46.7 & 68.6 & 0.314 & 10.8 & 45.3 & 68.5 & 0.126 & 0.0 & 18.1 & 35.0 \\
DE-DistMult & 0.501 & 39.2 & 56.9 & 70.8 & 0.484 & 36.6 & 54.6 & 71.8 & 0.213 & 13.0 & 22.8 & 37.6 \\
DE-SimplE & \textbf{0.526} & \textbf{41.8} & \textbf{59.2} & \textbf{72.5} & \textbf{0.513} & \textbf{39.2} & \textbf{57.8} & \textbf{74.8} & \textbf{0.230} & \textbf{14.1} & \textbf{24.8} & \textbf{40.3}
\end{tabular}
\end{center}
\end{table*}

\subsection{Model Variants \& Ablation Study} \label{sec:ablation}
\begin{figure}
   \centering
   \subfloat[]{%
   \includegraphics[width=0.35\textwidth]{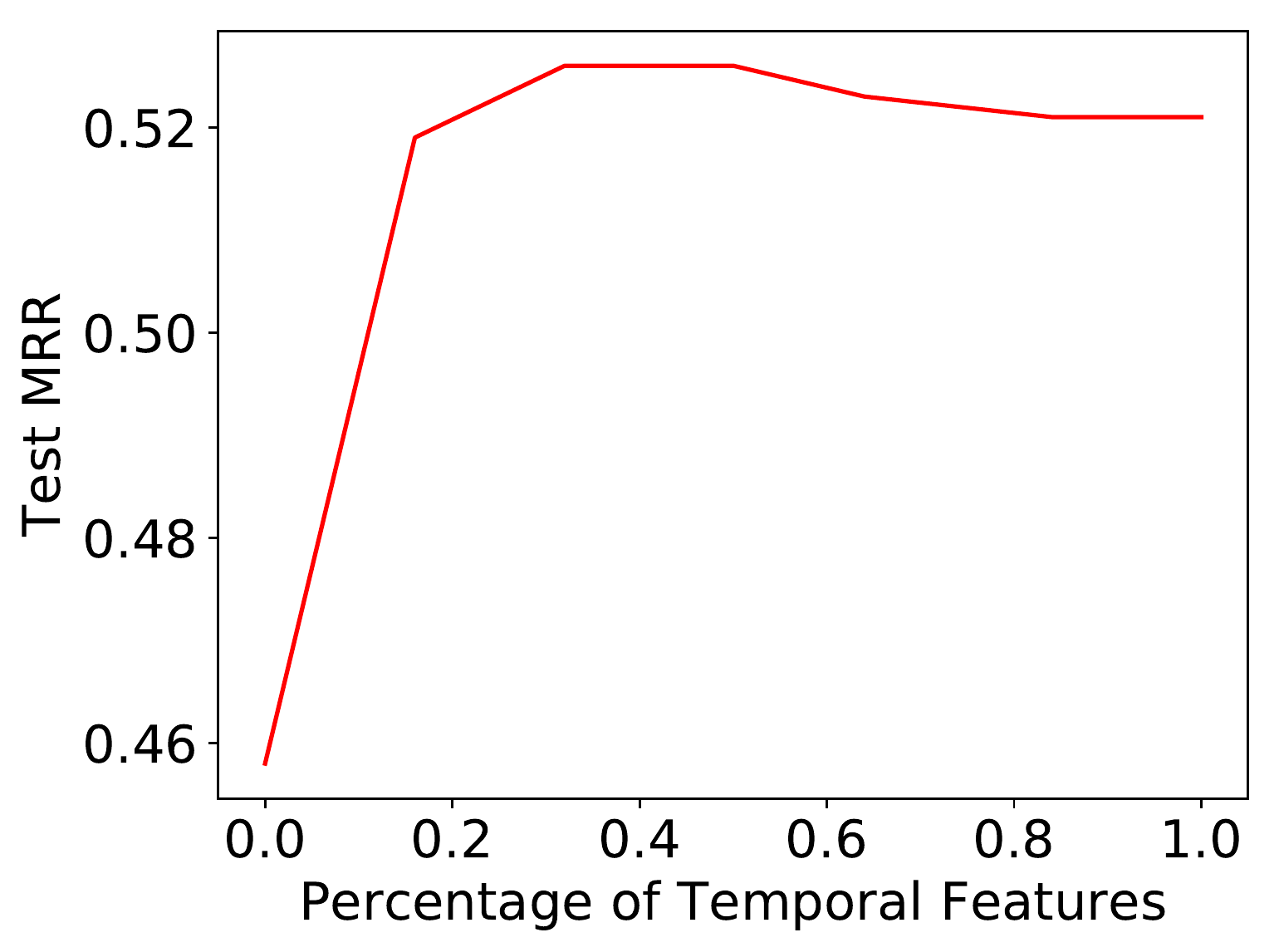}}
~~~~\hspace*{1cm}
   \subfloat[]{%
   \includegraphics[width=0.35\textwidth]{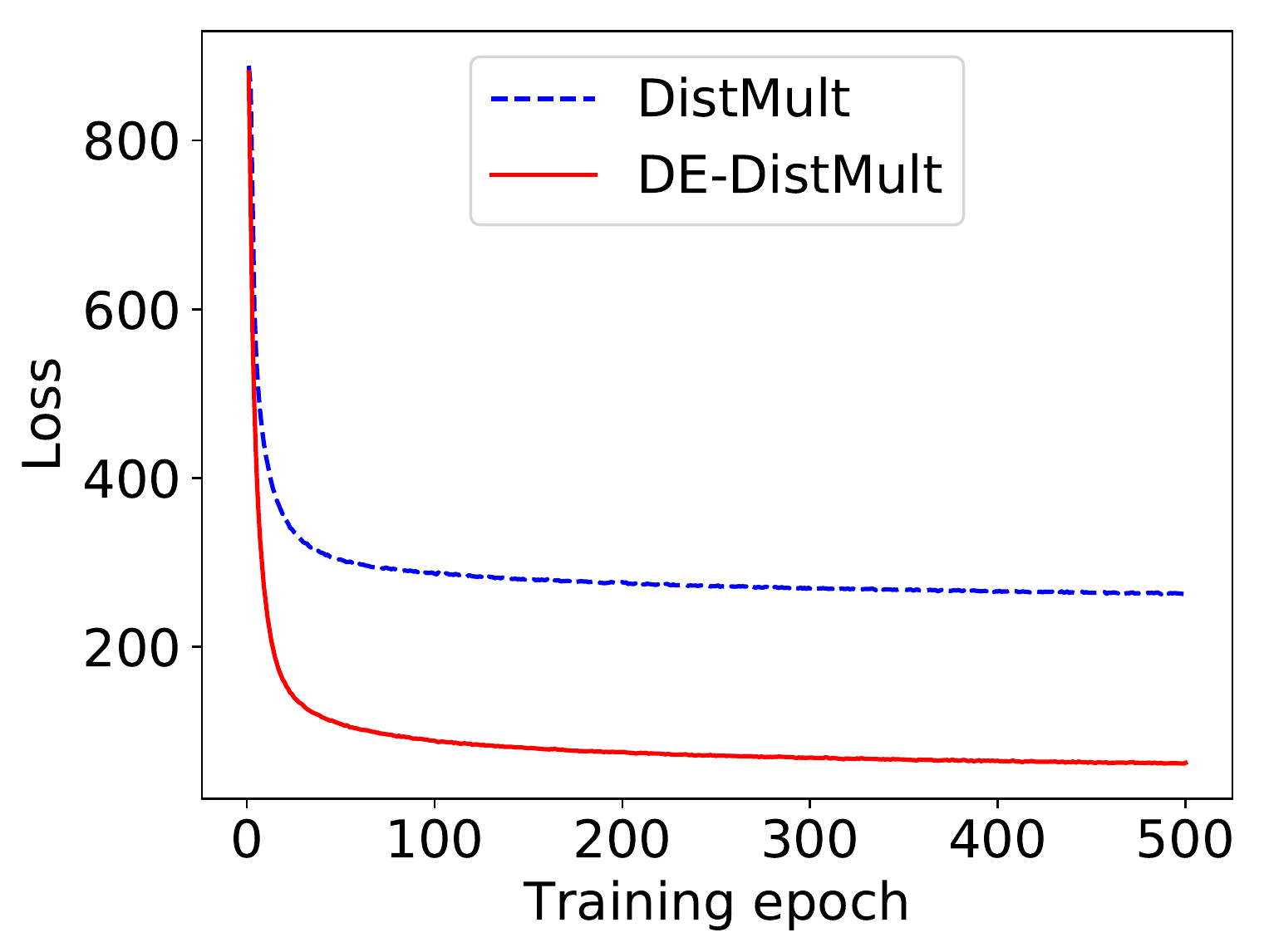}} %

   \caption{%
   \label{fig:static-training} %
   (a) Test MRR of DE-SimplE on ICEWS14 as a function of $\gamma$. (b) The training curve for DistMult and DE-DistMult.}
\end{figure}

We run experiments on ICEWS14 with several variants of the proposed models to provide a better understanding of them. The results can be found in Table~\ref{tab:variations} and Figure~\ref{fig:static-training}.  Table~\ref{tab:variations} includes DE-TransE and DE-DistMult with no variants as well so other variants can be easily compared to them.

\textbf{Activation Function:} So far, we used \emph{sine} as the activation function in Equation~\ref{eq:demb}. The performance for other activation functions including \emph{Tanh}, \emph{sigmoid}, \emph{Leaky ReLU} (with $0.1$ leakage), and \emph{squared exponential} are presented in Table~\ref{tab:variations}. From the table, it can be viewed that other activation functions also perform well. Specifically, squared exponential performs almost on-par with sine. 
We believe one reason why sine and squared exponential give better performance is because a combination of sine or square exponential features can generate more sophisticated features than a combination of Tanh, sigmoid, or ReLU features. While a temporal feature with Tanh or sigmoid as the activation corresponds to a smooth off-on (or on-off) temporal switch, a temporal feature with sine or squared exponential activation corresponds to two (or more) switches (e.g., off-on-off) which can potentially model relations that start at some time and end after a while (e.g., \relation{PresidentOf}). These results also provide evidence for the effectiveness of diachronic embedding across several $\function{DEEMB}$ functions.

\textbf{Adding Diachronic Embedding for Relations:} Compared to entities, we hypothesize that relations may evolve at a very lower rate or, for some relations, evolve only negligibly. Therefore, modeling relations with a static (rather than a diachronic) representation  may suffice. To test this hypothesis, we ran DE-TransE and DE-DistMult on ICEWS14 where relation embeddings are also a function of time. From the obtained results in Table~\ref{tab:variations}, one can see that the model with diachronic embeddings for both entities and relations performs on-par with the model with diachronic embedding only for entities. We conducted the same experiment on ICEWS05-15 (which has a longer time horizons) and GDELT and observed similar results. These results show that at least on our benchmarks, modeling the evolution of relations may not be helpful. Future work can test this hypothesis on datasets with other types of relations and longer horizons.

\textbf{Generalizing to Unseen Timestamps:} To measure how well our models generalize to timestamps not observed in train set, we created a variant of the ICEWS14 dataset by including every fact except those on the $\Th{5}$, $\Th{15}$, and $\Th{25}$ day of each month in the train set. We split the excluded facts randomly into validation and test sets (removing the ones including entities not observed in the train set). This ensures that none of the timestamps in the validation or test sets has been observed by the model in the train set. Then we ran DistMult and DE-DistMult on the resulting dataset. The obtained results in Table~\ref{tab:variations} indicate that DE-DistMult gains almost $10\%$ MRR improvement over DistMult thus showing the effectiveness of our diachronic embedding to generalize to unseen timestamps.

\textbf{Importance of Model Parameters Used in Equation~\ref{eq:demb}:} In Equation~\ref{eq:demb}, the temporal part of the embedding contains three  components: $\vctr{a}_\vertex{v}$, $\vctr{w}_\vertex{v}$, and $\vctr{b}_\vertex{v}$. To measure the importance of each component, we ran DE-DistMult on ICEWS14 under three settings: 1- when $\vctr{a}_\vertex{v}$s are removed (i.e. set to $1$), 2- when $\vctr{w}_\vertex{v}$s are removed (i.e. set to $1$), and 3- when $\vctr{b}_\vertex{v}$s are removed (i.e. set to $0$). From the obtained results presented in Table~\ref{tab:variations}, it can be viewed that all three components are important for the temporal features, especially $\vctr{a}_\vertex{v}$s and $\vctr{w}_\vertex{v}$s. 
Removing $\vctr{b}_\vertex{v}$s does not affect the results as much as $\vctr{a}_\vertex{v}$s and $\vctr{w}_\vertex{v}$s. 
Therefore, if one needs to reduce the number of parameters, removing $\vctr{b}_\vertex{v}$ may be a good option as long as they can tolerate a slight reduction in accuracy.

\begin{table}[t]
\setlength{\tabcolsep}{4pt}
\footnotesize
\caption{Results for different variations of our model on ICEWS14.}
\label{tab:variations}
\begin{center}
\begin{tabular}{c|c|c|c|c|c}
Model & Variation & MRR & Hit@1 & Hit@3 & Hit@10 \\ \hline
DE-TransE & No variation (Activation function: \emph{Sine}) & 0.326 & 12.4 & 46.7 & 68.6 \\
DE-DistMult & No variation (Activation function: \emph{Sine}) & 0.501 & 39.2 & 56.9 & 70.8 \\ \hline
DE-DistMult & Activation function: \emph{Tanh} & 0.486 & 37.5 & 54.7 & 70.1 \\
DE-DistMult & Activation function: \emph{Sigmoid} & 0.484 & 37.0 & 54.6 & 70.6 \\
DE-DistMult & Activation function: \emph{Leaky ReLU} & 0.478 & 36.3 & 54.2 & 70.1 \\
DE-DistMult & Activation function: \emph{Squared Exponential} & 0.501 & 39.0 & 56.8 & 70.9 \\ \hline
DE-TransE & Diachronic embedding for both entities and relations & 0.324 & 12.7 & 46.1 & 68.0 \\
DE-DistMult & Diachronic embedding for both entities and relations & 0.502 & 39.4 & 56.6 & 70.4 \\ \hline
DistMult & Generalizing to unseen timestamps & 0.410 & 30.2 & 46.2 & 62.0 \\
DE-DistMult & Generalizing to unseen timestamps & 0.452 & 34.5 & 51.3 & 65.4\\ \hline
DE-DistMult & $\,\vctr{a}_\vertex{v}[n]=1$~~for $1\leq n\leq \gamma d\:$ for all $\vertex{v}\in\vertices{V}$ & 0.458 & 34.4 & 51.8 & 68.3 \\
DE-DistMult & $\vctr{w}_\vertex{v}[n]=1$~~for $1\leq n\leq \gamma d\:$ for all $\vertex{v}\in\vertices{V}$ & 0.470 & 36.4 & 53.1 & 67.1 \\
DE-DistMult & $\:\vctr{b}_\vertex{v}[n]=0$~~for $1\leq n\leq \gamma d\:$ for all $\vertex{v}\in\vertices{V}$ & 0.498 & 38.9 & 56.2 & 70.4 \\
\end{tabular}
\end{center}
\end{table}

\textbf{Static Features:} Figure~\ref{fig:static-training}(a) shows the test MRR of DE-SimplE on ICEWS14 as a function of $\gamma$, the percentage of temporal features. According to Figure~\ref{fig:static-training}(a), as soon as some features become temporal (i.e. $\gamma$ changes from $0$ to a non-zero number), a substantial boost in performance can be observed. This observation sheds more light on the importance of learning temporal features and having diachronic embeddings. As $\gamma$ becomes larger, MRR reaches a peak and then slightly drops. This slight drop in performance can be due to overfitting to temporal cues. 
This result demonstrates that modeling static features explicitly can help reduce the number of learnable parameters and avoid overfitting. Such a design choice may be even more important when the embedding dimensions are larger. However, it comes at the cost of adding one hyper-parameter to the model. If one prefers a slightly less accurate model with fewer hyper-parameters, they can make all vector elements temporal.

\textbf{Training Curve:} Figure~\ref{fig:static-training}(b) shows the training curve for DistMult and DE-DistMult on ICEWS14. While it has been argued that using sine activation functions may complicate training in some neural network architectures (see, e.g., \cite{lapedes1987nonlinear,taming2017}), it can be viewed that when using sine activations, the training curve for our model is quite stable.  

\section{Related Work}\label{sec:related-work}
\textbf{StaRAI:} Statistical relational AI (StaRAI) \cite{raedt2016statistical,koller2007introduction} approaches are mainly based on soft (hanf-crafted or learned) rules \cite{richardson2006markov,de2007problog,kimmig2012short,kazemi2014relational} where the probability of a world is typically proportional to the number of rules that are satisfied/violated in that world and the confidence for each rule. A line of work in this area combines a stack of soft rules with embeddings for property prediction \cite{sourek2015lifted,kazemi2018relnn}. Another line of work extends the soft rules to temporal KGs \cite{sadilek2010recognizing,papai2012slice,dylla2013temporal,huber2014applying,chekol2017marrying,chekol2018rule}. The approaches based on soft rules have been generally shown to perform subpar to KG embedding models \cite{nickel2016review}.

\textbf{Graph Walk:} These approaches define weighted template walks on a KG and then answer queries by template matching \cite{lao2010relational,lao2011random}. They have been shown to be quite similar to, and in some cases subsumed by, the models based on soft rules \cite{kazemi2018bridging}.

\textbf{Static KG Embedding:} A large number of models have been developed for static KG embedding. A class of these models are the translational approaches corresponding to variations of TransE (see, e.g., \cite{lin2015learning,wang2014knowledge,StransE}). Another class of approaches are based on a bilinear score function $\transpose{\vctr{z}}_\vertex{v}\mtrx{Z}_\relation{r}\vctr{z}_\vertex{u}$ each imposing a different sparsity constraint on the $\mtrx{Z}_\relation{r}$ matrices (see, e.g., \cite{nickel2011three,trouillon2016complex,nickel2016holographic,kazemi2018simple,liu2017analogical}). A third class of models are based on deep learning approaches using feed-forward or convolutional layers on top of the embeddings (see, e.g., \cite{socher2013reasoning,dong2014knowledge,dettmers2018convolutional,balazevic2018hypernetwork}). These models can be potentially extended to TKGC through our diachronic embedding.

\textbf{Temporal KG Embedding:} Several works have extended the static KG embedding models to temporal KGs. \citet{jiang2016towards} extend TransE by adding atimestamp embedding into the score function. \citet{dasgupta2018hyte} extend TransE by projecting the embeddings to the timestamp hyperplain and then using the TransE score on the projected space. \citet{ma2018embedding} extend several models by adding a timestamp embedding to their score functions. These models may not work well when the number of timestamps is large. Furthermore, since they only learn embeddings for observed timestamps, they cannot generalize to unseen timestamps. \citet{garcia2018learning} extend TransE and DistMult by combining the relation and timestamp through a character LSTM. 
These models have been described in detail in Section~\ref{sec:background} and their performances have been reported in Table~\ref{tab:results}.

\textbf{KG Embedding for Extrapolation:} TKGC is an \emph{interpolation} problem where given a set of temporal facts in a time frame, the goal is to predict the missing facts. A related problem is the \emph{extrapolation} problem where future interactions are to be predicted (see, e.g., \cite{trivedi2017know,kumar2018learning,trivedi2019dyrep}). Despite some similarities in the employed approaches, KG extrapolation is fundamentally different from TKGC in that a score for an interaction $(\vertex{v}, \relation{r}, \vertex{u}, \timestamp{t})$ is to be computed given only the past (i.e. facts before $\timestamp{t}$) whereas in TKGC the score is to be computed given past, present, and future. A comprehensive analysis of the existing models for interpolation and extrapolation can be found in \cite{kazemi2019relational}.

\textbf{Diachronic Word Embeddings:} The idea behind our proposed embeddings is similar to diachronic word embeddings where a corpus is typically broken temporally into slices (e.g., 20-year chuncks of a 200-year corpus) and embeddings are learned for words in each chunk thus providing word embeddings that are a function of time (see, e.g., \cite{kim2014temporal,kulkarni2015statistically,hamilton2016diachronic,bamler2017dynamic}). The main goal of diachronic word embeddings is to reveal how the meanings of the words have evolved over time. Our work can be viewed as an extension of diachronic word embeddings to continuous-time KG completion.

\section{Conclusion}
Temporal knowledge graph (KG) completion is an important problem and has been the focus of several recent studies. We developed a diachronic embedding function for temporal KG completion which provides a hidden representation for the entities of a temporal KG at any point in time. Our embedding is generic and can be combined with any score function. We proved that combining our diachronic embedding with SimplE results in a fully expressive model -- the first temporal KG embedding model for which such a result exists. We showed the superior performance of our model compared to existing work on several benchmarks. Future work includes designing functions other than the one proposed in Equation~\ref{eq:demb}, a comprehensive study of which functions are favored by different types of KGs, and using our proposed embedding for diachronic word embedding. 

\bibliography{MyBib}

\begin{thebibliography}{}

\bibitem[\protect\citeauthoryear{Balazevic \bgroup \em et al.\egroup
  }{2018}]{balazevic2018hypernetwork}
Ivana Balazevic, Carl Allen, and Timothy~M Hospedales.
\newblock Hypernetwork knowledge graph embeddings.
\newblock {\em arXiv preprint arXiv:1808.07018}, 2018.

\bibitem[\protect\citeauthoryear{Bala{\v{z}}evi{\'c} \bgroup \em et al.\egroup
  }{2019}]{balavzevic2019tucker}
Ivana Bala{\v{z}}evi{\'c}, Carl Allen, and Timothy~M Hospedales.
\newblock Tucker: Tensor factorization for knowledge graph completion.
\newblock {\em arXiv preprint arXiv:1901.09590}, 2019.

\bibitem[\protect\citeauthoryear{Bamler and Mandt}{2017}]{bamler2017dynamic}
Robert Bamler and Stephan Mandt.
\newblock Dynamic word embeddings.
\newblock In {\em ICML}, pages 380--389, 2017.

\bibitem[\protect\citeauthoryear{Bordes \bgroup \em et al.\egroup
  }{2013}]{bordes2013translating}
Antoine Bordes, Nicolas Usunier, Alberto Garcia-Duran, Jason Weston, and Oksana
  Yakhnenko.
\newblock Translating embeddings for modeling multi-relational data.
\newblock In {\em NeurIPS}, pages 2787--2795, 2013.

\bibitem[\protect\citeauthoryear{Boschee \bgroup \em et al.\egroup
  }{2015}]{boschee2015icews}
Elizabeth Boschee, Jennifer Lautenschlager, Sean O’Brien, Steve Shellman,
  James Starz, and Michael Ward.
\newblock Icews coded event data.
\newblock {\em Harvard Dataverse}, 12, 2015.

\bibitem[\protect\citeauthoryear{Buchman and Poole}{2016}]{buchman2016negation}
David Buchman and David Poole.
\newblock Negation without negation in probabilistic logic programming.
\newblock In {\em KR}, 2016.

\bibitem[\protect\citeauthoryear{Carslaw}{1921}]{carslaw1921introduction}
Horatio~Scott Carslaw.
\newblock {\em Introduction to the Theory of Fourier's Series and Integrals}.
\newblock Macmillan, 1921.

\bibitem[\protect\citeauthoryear{Chekol and
  Stuckenschmidt}{2018}]{chekol2018rule}
Melisachew~Wudage Chekol and Heiner Stuckenschmidt.
\newblock Rule based temporal inference.
\newblock In {\em ICLP}. Schloss Dagstuhl-Leibniz-Zentrum fuer Informatik,
  2018.

\bibitem[\protect\citeauthoryear{Chekol \bgroup \em et al.\egroup
  }{2017}]{chekol2017marrying}
Melisachew~Wudage Chekol, Giuseppe Pirr{\`o}, Joerg Schoenfisch, and Heiner
  Stuckenschmidt.
\newblock Marrying uncertainty and time in knowledge graphs.
\newblock In {\em AAAI}, 2017.

\bibitem[\protect\citeauthoryear{Dasgupta \bgroup \em et al.\egroup
  }{2018}]{dasgupta2018hyte}
Shib~Sankar Dasgupta, Swayambhu~Nath Ray, and Partha Talukdar.
\newblock Hyte: Hyperplane-based temporally aware knowledge graph embedding.
\newblock In {\em EMNLP}, pages 2001--2011, 2018.

\bibitem[\protect\citeauthoryear{De~Raedt \bgroup \em et al.\egroup
  }{2007}]{de2007problog}
Luc De~Raedt, Angelika Kimmig, and Hannu Toivonen.
\newblock Problog: A probabilistic prolog and its application in link
  discovery.
\newblock In {\em IJCAI}, volume~7, pages 2462--2467. Hyderabad, 2007.

\bibitem[\protect\citeauthoryear{Dettmers \bgroup \em et al.\egroup
  }{2018}]{dettmers2018convolutional}
Tim Dettmers, Pasquale Minervini, Pontus Stenetorp, and Sebastian Riedel.
\newblock Convolutional 2d knowledge graph embeddings.
\newblock In {\em AAAI}, 2018.

\bibitem[\protect\citeauthoryear{Dong \bgroup \em et al.\egroup
  }{2014}]{dong2014knowledge}
Xin Dong, Evgeniy Gabrilovich, Geremy Heitz, Wilko Horn, Ni~Lao, Kevin Murphy,
  Thomas Strohmann, Shaohua Sun, and Wei Zhang.
\newblock Knowledge vault: A web-scale approach to probabilistic knowledge
  fusion.
\newblock In {\em ACM SIGKDD}, pages 601--610. ACM, 2014.

\bibitem[\protect\citeauthoryear{Dylla \bgroup \em et al.\egroup
  }{2013}]{dylla2013temporal}
Maximilian Dylla, Iris Miliaraki, and Martin Theobald.
\newblock A temporal-probabilistic database model for information extraction.
\newblock {\em Proceedings of the VLDB Endowment}, 6(14):1810--1821, 2013.

\bibitem[\protect\citeauthoryear{Fatemi \bgroup \em et al.\egroup
  }{2019}]{fatemi2019improved}
Bahare Fatemi, Siamak Ravanbakhsh, and David Poole.
\newblock Improved knowledge graph embedding using background taxonomic
  information.
\newblock In {\em AAAI}, 2019.

\bibitem[\protect\citeauthoryear{Garc{\'\i}a-Dur{\'a}n \bgroup \em et
  al.\egroup }{2018}]{garcia2018learning}
Alberto Garc{\'\i}a-Dur{\'a}n, Sebastijan Duman{\v{c}}i{\'c}, and Mathias
  Niepert.
\newblock Learning sequence encoders for temporal knowledge graph completion.
\newblock {\em arXiv preprint arXiv:1809.03202}, 2018.

\bibitem[\protect\citeauthoryear{Giambattista~Parascandolo}{2017}]{taming2017}
Tuomas~Virtanen Giambattista~Parascandolo, Heikki~Huttunen.
\newblock Taming the waves: sine as activation function in deep neural
  networks.
\newblock 2017.

\bibitem[\protect\citeauthoryear{Hamilton \bgroup \em et al.\egroup
  }{2016}]{hamilton2016diachronic}
William~L Hamilton, Jure Leskovec, and Dan Jurafsky.
\newblock Diachronic word embeddings reveal statistical laws of semantic
  change.
\newblock {\em arXiv preprint arXiv:1605.09096}, 2016.

\bibitem[\protect\citeauthoryear{Hitchcock}{1927}]{hitchcock1927expression}
Frank~L Hitchcock.
\newblock The expression of a tensor or a polyadic as a sum of products.
\newblock {\em Journal of Mathematics and Physics}, 6(1-4):164--189, 1927.

\bibitem[\protect\citeauthoryear{Huber \bgroup \em et al.\egroup
  }{2014}]{huber2014applying}
Jakob Huber, Christian Meilicke, and Heiner Stuckenschmidt.
\newblock Applying {M}arkov logic for debugging probabilistic temporal
  knowledge bases.
\newblock In {\em AKBC}, 2014.

\bibitem[\protect\citeauthoryear{Jiang \bgroup \em et al.\egroup
  }{2016}]{jiang2016towards}
Tingsong Jiang, Tianyu Liu, Tao Ge, Lei Sha, Baobao Chang, Sujian Li, and
  Zhifang Sui.
\newblock Towards time-aware knowledge graph completion.
\newblock In {\em COLING}, pages 1715--1724, 2016.

\bibitem[\protect\citeauthoryear{Kadlec and
  Kleindienst}{2017}]{kadlec2017knowledge}
Ondrej~Bajgar Kadlec, Rudolf and Jan Kleindienst.
\newblock Knowledge base completion: Baselines strike back.
\newblock {\em arXiv preprint arXiv:1705.10744}, 2017.

\bibitem[\protect\citeauthoryear{Kazemi and Poole}{2018a}]{kazemi2018bridging}
Seyed~Mehran Kazemi and David Poole.
\newblock Bridging weighted rules and graph random walks for statistical
  relational models.
\newblock {\em Frontiers in Robotics and AI}, 5:8, 2018.

\bibitem[\protect\citeauthoryear{Kazemi and Poole}{2018b}]{kazemi2018relnn}
Seyed~Mehran Kazemi and David Poole.
\newblock Rel{NN}: A deep neural model for relational learning.
\newblock In {\em AAAI}, 2018.

\bibitem[\protect\citeauthoryear{Kazemi and Poole}{2018c}]{kazemi2018simple}
Seyed~Mehran Kazemi and David Poole.
\newblock Simpl{E} embedding for link prediction in knowledge graphs.
\newblock In {\em NeurIPS}, pages 4289--4300, 2018.

\bibitem[\protect\citeauthoryear{Kazemi \bgroup \em et al.\egroup
  }{2014}]{kazemi2014relational}
Seyed~Mehran Kazemi, David Buchman, Kristian Kersting, Sriraam Natarajan, and
  David Poole.
\newblock Relational logistic regression.
\newblock In {\em KR}, 2014.

\bibitem[\protect\citeauthoryear{Kazemi \bgroup \em et al.\egroup
  }{2019}]{kazemi2019relational}
Seyed~Mehran Kazemi, Rishab Goel, Kshitij Jain, Ivan Kobyzev, Akshay Sethi,
  Peter Forsyth, and Pascal Poupart.
\newblock Relational representation learning for dynamic (knowledge) graphs: A
  survey.
\newblock {\em arXiv preprint arXiv:1905.11485}, 2019.

\bibitem[\protect\citeauthoryear{Kim \bgroup \em et al.\egroup
  }{2014}]{kim2014temporal}
Yoon Kim, Yi-I Chiu, Kentaro Hanaki, Darshan Hegde, and Slav Petrov.
\newblock Temporal analysis of language through neural language models.
\newblock {\em arXiv preprint arXiv:1405.3515}, 2014.

\bibitem[\protect\citeauthoryear{Kimmig \bgroup \em et al.\egroup
  }{2012}]{kimmig2012short}
Angelika Kimmig, Stephen~H Bach, Matthias Broecheler, Bert Huang, and Lise
  Getoor.
\newblock A short introduction to probabilistic soft logic.
\newblock In {\em NIPS Workshop on probabilistic programming: Foundations and
  applications}, volume~1, page~3, 2012.

\bibitem[\protect\citeauthoryear{Kingma and Ba}{2014}]{kingma2014adam}
Diederik~P Kingma and Jimmy Ba.
\newblock Adam: A method for stochastic optimization.
\newblock {\em arXiv preprint arXiv:1412.6980}, 2014.

\bibitem[\protect\citeauthoryear{Koller \bgroup \em et al.\egroup
  }{2007}]{koller2007introduction}
Daphne Koller, Nir Friedman, Sa{\v{s}}o D{\v{z}}eroski, Charles Sutton, Andrew
  McCallum, Avi Pfeffer, Pieter Abbeel, Ming-Fai Wong, David Heckerman, Chris
  Meek, et~al.
\newblock {\em Introduction to statistical relational learning}.
\newblock MIT press, 2007.

\bibitem[\protect\citeauthoryear{Kulkarni \bgroup \em et al.\egroup
  }{2015}]{kulkarni2015statistically}
Vivek Kulkarni, Rami Al-Rfou, Bryan Perozzi, and Steven Skiena.
\newblock Statistically significant detection of linguistic change.
\newblock In {\em WWW}, pages 625--635, 2015.

\bibitem[\protect\citeauthoryear{Kumar \bgroup \em et al.\egroup
  }{2018}]{kumar2018learning}
Srijan Kumar, Xikun Zhang, and Jure Leskovec.
\newblock Learning dynamic embedding from temporal interaction networks.
\newblock {\em arXiv preprint arXiv:1812.02289}, 2018.

\bibitem[\protect\citeauthoryear{Lacroix \bgroup \em et al.\egroup
  }{2018}]{lacroix2018canonical}
Timoth{\'e}e Lacroix, Nicolas Usunier, and Guillaume Obozinski.
\newblock Canonical tensor decomposition for knowledge base completion.
\newblock In {\em ICML}, 2018.

\bibitem[\protect\citeauthoryear{Lao and Cohen}{2010}]{lao2010relational}
Ni~Lao and William~W Cohen.
\newblock Relational retrieval using a combination of path-constrained random
  walks.
\newblock {\em Machine learning}, 81(1):53--67, 2010.

\bibitem[\protect\citeauthoryear{Lao \bgroup \em et al.\egroup
  }{2011}]{lao2011random}
Ni~Lao, Tom Mitchell, and William~W Cohen.
\newblock Random walk inference and learning in a large scale knowledge base.
\newblock In {\em EMNLP}, pages 529--539, 2011.

\bibitem[\protect\citeauthoryear{Lapedes and
  Farber}{1987}]{lapedes1987nonlinear}
Alan Lapedes and Robert Farber.
\newblock Nonlinear signal processing using neural networks: Prediction and
  system modelling.
\newblock Technical report, 1987.

\bibitem[\protect\citeauthoryear{Leetaru and Schrodt}{2013}]{leetaru2013gdelt}
Kalev Leetaru and Philip~A Schrodt.
\newblock Gdelt: Global data on events, location, and tone, 1979--2012.
\newblock In {\em ISA annual convention}, volume~2, pages 1--49. Citeseer,
  2013.

\bibitem[\protect\citeauthoryear{Lin \bgroup \em et al.\egroup
  }{2015}]{lin2015learning}
Yankai Lin, Zhiyuan Liu, Maosong Sun, Yang Liu, and Xuan Zhu.
\newblock Learning entity and relation embeddings for knowledge graph
  completion.
\newblock In {\em AAAI}, pages 2181--2187, 2015.

\bibitem[\protect\citeauthoryear{Liu \bgroup \em et al.\egroup
  }{2017}]{liu2017analogical}
Hanxiao Liu, Yuexin Wu, and Yiming Yang.
\newblock Analogical inference for multi-relational embeddings.
\newblock In {\em ICML}, pages 2168--2178, 2017.

\bibitem[\protect\citeauthoryear{Ma \bgroup \em et al.\egroup
  }{2018}]{ma2018embedding}
Yunpu Ma, Volker Tresp, and Erik~A Daxberger.
\newblock Embedding models for episodic knowledge graphs.
\newblock {\em Journal of Web Semantics}, 2018.

\bibitem[\protect\citeauthoryear{Minervini \bgroup \em et al.\egroup
  }{2017}]{minervini2017regularizing}
Pasquale Minervini, Luca Costabello, Emir Mu{\~n}oz, V{\'\i}t
  Nov{\'a}{\v{c}}ek, and Pierre-Yves Vandenbussche.
\newblock Regularizing knowledge graph embeddings via equivalence and inversion
  axioms.
\newblock In {\em ECML PKDD}, pages 668--683. Springer, 2017.

\bibitem[\protect\citeauthoryear{Nguyen \bgroup \em et al.\egroup
  }{2016}]{StransE}
Dat~Quoc Nguyen, Kairit Sirts, Lizhen Qu, and Mark Johnson.
\newblock Stranse: a novel embedding model of entities and relationships in
  knowledge bases.
\newblock In {\em NAACL-HLT}, 2016.

\bibitem[\protect\citeauthoryear{Nguyen}{2017}]{nguyen2017overview}
Dat~Quoc Nguyen.
\newblock An overview of embedding models of entities and relationships for
  knowledge base completion.
\newblock {\em arXiv preprint arXiv:1703.08098}, 2017.

\bibitem[\protect\citeauthoryear{Nickel \bgroup \em et al.\egroup
  }{2011}]{nickel2011three}
Maximilian Nickel, Volker Tresp, and Hans-Peter Kriegel.
\newblock A three-way model for collective learning on multi-relational data.
\newblock In {\em ICML}, volume~11, pages 809--816, 2011.

\bibitem[\protect\citeauthoryear{Nickel \bgroup \em et al.\egroup
  }{2016a}]{nickel2016review}
Maximilian Nickel, Kevin Murphy, Volker Tresp, and Evgeniy Gabrilovich.
\newblock A review of relational machine learning for knowledge graphs.
\newblock {\em Proceedings of the IEEE}, 104(1):11--33, 2016.

\bibitem[\protect\citeauthoryear{Nickel \bgroup \em et al.\egroup
  }{2016b}]{nickel2016holographic}
Maximilian Nickel, Lorenzo Rosasco, and Tomaso Poggio.
\newblock Holographic embeddings of knowledge graphs.
\newblock In {\em AAAI}, 2016.

\bibitem[\protect\citeauthoryear{Papai \bgroup \em et al.\egroup
  }{2012}]{papai2012slice}
Tivadar Papai, Henry Kautz, and Daniel Stefankovic.
\newblock Slice normalized dynamic markov logic networks.
\newblock In {\em NeurIPS}, pages 1907--1915, 2012.

\bibitem[\protect\citeauthoryear{Paszke \bgroup \em et al.\egroup
  }{2017}]{paszke2017automatic}
Adam Paszke, Sam Gross, Soumith Chintala, Gregory Chanan, Edward Yang, Zachary
  DeVito, Zeming Lin, Alban Desmaison, Luca Antiga, and Adam Lerer.
\newblock Automatic differentiation in pytorch.
\newblock In {\em NIPS-W}, 2017.

\bibitem[\protect\citeauthoryear{Raedt \bgroup \em et al.\egroup
  }{2016}]{raedt2016statistical}
Luc~De Raedt, Kristian Kersting, Sriraam Natarajan, and David Poole.
\newblock Statistical relational artificial intelligence: Logic, probability,
  and computation.
\newblock {\em Synthesis Lectures on Artificial Intelligence and Machine
  Learning}, 10(2):1--189, 2016.

\bibitem[\protect\citeauthoryear{Richardson and
  Domingos}{2006}]{richardson2006markov}
Matthew Richardson and Pedro Domingos.
\newblock Markov logic networks.
\newblock {\em Machine learning}, 62(1-2):107--136, 2006.

\bibitem[\protect\citeauthoryear{Sadilek and
  Kautz}{2010}]{sadilek2010recognizing}
Adam Sadilek and Henry Kautz.
\newblock Recognizing multi-agent activities from gps data.
\newblock In {\em AAAI}, 2010.

\bibitem[\protect\citeauthoryear{Socher \bgroup \em et al.\egroup
  }{2013}]{socher2013reasoning}
Richard Socher, Danqi Chen, Christopher~D Manning, and Andrew Ng.
\newblock Reasoning with neural tensor networks for knowledge base completion.
\newblock In {\em AAAI}, pages 926--934, 2013.

\bibitem[\protect\citeauthoryear{Sourek \bgroup \em et al.\egroup
  }{2015}]{sourek2015lifted}
Gustav Sourek, Vojtech Aschenbrenner, Filip Zelezny, and Ondrej Kuzelka.
\newblock Lifted relational neural networks.
\newblock {\em arXiv preprint arXiv:1508.05128}, 2015.

\bibitem[\protect\citeauthoryear{Sun \bgroup \em et al.\egroup
  }{2019}]{sun2019rotate}
Zhiqing Sun, Zhi-Hong Deng, Jian-Yun Nie, and Jian Tang.
\newblock Rotat{E}: Knowledge graph embedding by relational rotation in complex
  space.
\newblock In {\em ICLR}, 2019.

\bibitem[\protect\citeauthoryear{Trivedi \bgroup \em et al.\egroup
  }{2017}]{trivedi2017know}
Rakshit Trivedi, Hanjun Dai, Yichen Wang, and Le~Song.
\newblock Know-evolve: Deep temporal reasoning for dynamic knowledge graphs.
\newblock In {\em ICML}, pages 3462--3471, 2017.

\bibitem[\protect\citeauthoryear{Trivedi \bgroup \em et al.\egroup
  }{2019}]{trivedi2019dyrep}
Rakshit Trivedi, Mehrdad Farajtabar, Prasenjeet Biswal, and Hongyuan Zha.
\newblock Dy{R}ep: Learning representations over dynamic graphs.
\newblock In {\em ICLR}, 2019.

\bibitem[\protect\citeauthoryear{Trouillon \bgroup \em et al.\egroup
  }{2016}]{trouillon2016complex}
Th{\'e}o Trouillon, Johannes Welbl, Sebastian Riedel, {\'E}ric Gaussier, and
  Guillaume Bouchard.
\newblock Complex embeddings for simple link prediction.
\newblock In {\em ICML}, pages 2071--2080, 2016.

\bibitem[\protect\citeauthoryear{Trouillon \bgroup \em et al.\egroup
  }{2017}]{trouillon2017knowledge}
Th{\'e}o Trouillon, Christopher~R Dance, {\'E}ric Gaussier, Johannes Welbl,
  Sebastian Riedel, and Guillaume Bouchard.
\newblock Knowledge graph completion via complex tensor factorization.
\newblock {\em JMLR}, 18(1):4735--4772, 2017.

\bibitem[\protect\citeauthoryear{Tucker}{1966}]{tucker1966some}
Ledyard~R Tucker.
\newblock Some mathematical notes on three-mode factor analysis.
\newblock {\em Psychometrika}, 31(3):279--311, 1966.

\bibitem[\protect\citeauthoryear{Wang \bgroup \em et al.\egroup
  }{2014}]{wang2014knowledge}
Zhen Wang, Jianwen Zhang, Jianlin Feng, and Zheng Chen.
\newblock Knowledge graph embedding by translating on hyperplanes.
\newblock In {\em AAAI}, 2014.

\bibitem[\protect\citeauthoryear{Wang \bgroup \em et al.\egroup
  }{2017}]{wang2017knowledge}
Quan Wang, Zhendong Mao, Bin Wang, and Li~Guo.
\newblock Knowledge graph embedding: A survey of approaches and applications.
\newblock {\em IEEE TKDE}, 29(12):2724--2743, 2017.

\bibitem[\protect\citeauthoryear{Xu \bgroup \em et al.\egroup
  }{2019}]{xu2019powerful}
Keyulu Xu, Weihua Hu, Jure Leskovec, and Stefanie Jegelka.
\newblock How powerful are graph neural networks?
\newblock In {\em ICLR}, 2019.

\bibitem[\protect\citeauthoryear{Yang \bgroup \em et al.\egroup
  }{2015}]{yang2015embedding}
Bishan Yang, Wen-tau Yih, Xiaodong He, Jianfeng Gao, and Li~Deng.
\newblock Embedding entities and relations for learning and inference in
  knowledge bases.
\newblock {\em ICLR}, 2015.

\end{thebibliography}
\bibliographystyle{named}

\appendix
\section{Proof of Theorems and Propositions}
\label{appendix:proofs}
\setcounter{theorem}{0}
\setcounter{proposition}{0}

\begin{theorem}
DE-SimplE is fully expressive for temporal knowledge graph completion.
\end{theorem}

\begin{proof}
For every entity $\vertex{v}_i\in\vertices{V}$, let $\function{DEEMB}(\vertex{v}_i, t)=(\vec{\vctr{z}}^\timestamp{t}_{\vertex{v}_i}, \cev{\vctr{z}}^\timestamp{t}_{\vertex{v}_i})$ where, according to Equation~\ref{eq:demb} with sine activations, $\vec{\vctr{z}}^\timestamp{\:t}_{\vertex{v}_i}\in\mathbb{R}^d$ and $\cev{\vctr{z}}^\timestamp{\,t}_{\vertex{v}_i}\in\mathbb{R}^d$ are defined as follows:
\begin{equation}
  \vec{\vctr{z}}^\timestamp{\:t}_{\vertex{v}_i}[n]=\begin{cases}
    \vec{\vctr{a}}_{\vertex{v}_i}[n] \sin(\vec{\vctr{w}}_{\vertex{v}_i}[n] t + \vec{\vctr{b}}_{\vertex{v}_i}[n]), & \text{if~~$n\leq \gamma$}. \\
    \vec{\vctr{a}}_{\vertex{v}_i}[n], & \text{if~~$n > \gamma$}.
  \end{cases}
\end{equation}
and:
\begin{equation}
  \cev{\vctr{z}}^\timestamp{\,t}_{\vertex{v}_i}[n]=\begin{cases}
    \cev{\vctr{a}}_{\vertex{v}_i}[n] \sin(\cev{\vctr{w}}_{\vertex{v}_i}[n] t + \cev{\vctr{b}}_{\vertex{v}_i}[n]), & \text{if~~$n\leq \gamma$}. \\
    \cev{\vctr{a}}_{\vertex{v}_i}[n], & \text{if~~$n > \gamma$}.
  \end{cases}
\end{equation}
We provide the proof for a specific case of DE-SimplE where the elements of $\vec{\vctr{z}}^\timestamp{\:t}_\vertex{v}$s are all temporal and the elements of $\cev{\vctr{z}}^\timestamp{\,t}_\vertex{v}$s are all non-temporal. This specific case can be achieved by setting $\gamma=d$, and $\cev{\vctr{w}}_\vertex{v}[n]=0$ and $\cev{\vctr{b}}_\vertex{v}[n]=\frac{\pi}{2}$ for all $\vertex{v}\in\vertices{V}$ and for all $1\leq n \leq d$. If this specific case of DE-SimplE is fully expressive, so is DE-SimplE. In this specific case, $\vec{\vctr{z}}^\timestamp{t}_{\vertex{v}_i}$ and $\cev{\vctr{z}}^\timestamp{t}_{\vertex{v}_i}$ for every $\vertex{v}_i\in\vertices{V}$ can be re-written as follows:
\begin{align}
    \vec{\vctr{z}}^\timestamp{\:t}_{\vertex{v}_i}[n] =&~ \vec{\vctr{a}}_{\vertex{v}_i}[n] \sin(\vec{\vctr{w}}_{\vertex{v}_i}[n] t + \vec{\vctr{b}}_{\vertex{v}_i}[n]) \\
    \cev{\vctr{z}}^\timestamp{\,t}_{\vertex{v}_i}[n] =&~ \cev{\vctr{a}}_{\vertex{v}_i}[n]
\end{align}
For every relation $\relation{r}_j\in\relations{R}$, let $\function{REMB}(\relation{r})=(\vec{\vctr{z}}_{\relation{r}_j}, \cev{\vctr{z}}_{\relation{r}_j})$. To further simplify the proof, following \cite{kazemi2018simple}, we only show how the embedding values can be set such that $\sumElemProd{\vec{\vctr{z}}^\timestamp{\:t}_{\vertex{v}_i}, \vec{\vctr{z}}_{\relation{r}_j}, \cev{\vctr{z}}^\timestamp{\,t}_{\vertex{v}_k}}$ becomes a positive number if $(\vertex{v}_i, \relation{r}_j, \vertex{v}_k, \timestamp{t}) \in \world{W}$ and a negative number if $(\vertex{v}_i, \relation{r}_j, \vertex{v}_k, \timestamp{t}) \in \world{W}^c$. Extending the proof the case where the score contains both components ($\sumElemProd{\vec{\vctr{z}}^\timestamp{\:t}_{\vertex{v}_i}, \vec{\vctr{z}}_{\relation{r}_j}, \cev{\vctr{z}}^\timestamp{\,t}_{\vertex{v}_k}}$ and $\sumElemProd{\vec{\vctr{z}}^\timestamp{\:t}_{\vertex{v}_k}, \cev{\vctr{z}}_{\relation{r}_j}, \cev{\vctr{z}}^\timestamp{\,t}_{\vertex{v}_i}}$) can be done by doubling the size of the embedding vectors and following a similar procedure as the one explained below for the second half of the vectors.

Assume $d=|\relations{R}|\cdot|\vertices{V}|\cdot|\timestamps{T}|\cdot L$ where $L\in\mathbb{N}$ is a natural number. These vectors can be viewed as $|\relations{R}|$ blocks of size $|\vertices{V}|\cdot|\timestamps{T}|\cdot L$. For the $\Th{j}$ relation $\relation{r}_j$, let $\vec{\vctr{z}}_{\relation{r}_j}$ be zero everywhere except on the $\Th{j}$ block where it is $1$ everywhere. With such a value assignment to $\vec{\vctr{z}}_{\relation{r}_j}$s, to find the score for a fact $(\vertex{v}_i, \relation{r}_j, \vertex{v}_k, \timestamp{t})$, only the $\Th{j}$ block of each embedding vector is important. Let us now focus on the $\Th{j}$ block.

The size of the $\Th{j}$ block (similar to all other blocks) is $|\vertices{V}|\cdot|\timestamps{T}|\cdot L$ and it can be viewed as $|\vertices{V}|$ sub-blocks of size $|\timestamps{T}|\cdot L$. For the $\Th{i}$ entity $\vertex{v}_i$, let the values of $\vec{\vctr{a}}_{\vertex{v}_i}$ be zero in all sub-blocks except the $\Th{i}$ sub-block. With such a value assignment, to find the score for a fact $(\vertex{v}_i, \relation{r}_j, \vertex{v}_k, \timestamp{t})$, only the $\Th{i}$ sub-block of the $\Th{j}$ block is important. Note that this sub-block is unique for each tuple $(\vertex{v}_i, \relation{r}_j)$. Let us now focus on the $\Th{i}$ sub-block of the $\Th{j}$ block.

The size of the $\Th{i}$ sub-block of the $\Th{j}$ block is $|\timestamps{T}| \cdot L$ and it can be viewed as $|\timestamps{T}|$ sub-sub-blocks of size $L$. According to the Fourier sine series \cite{carslaw1921introduction}, with a large enough $L$, we can set the values for $\vec{\vctr{a}}_{\vertex{v}_i}$, $\vec{\vctr{w}}_{\vertex{v}_i}$, and $\vec{\vctr{b}}_{\vertex{v}_i}$ in a way that the sum of the elements of $\vec{\vctr{z}}^\timestamp{\:t}_{\vertex{v}_i}$ for the $\Th{p}$ sub-sub-block becomes $1$ when $\timestamp{t}= \timestamp{t}_p$ (where $\timestamp{t}_p$ is the $\Th{p}$ timestamp in $\timestamps{T}$) and $0$ when $t$ is a timestamp other than $\timestamp{t}_p$. Note that this sub-sub-block is unique for each tuple $(\vertex{v}_i, \relation{r}_j, \timestamp{t}_p)$.

Having the above value assignments, if $(\vertex{v}_i, \relation{r}_j, \vertex{v}_k, \timestamp{t}_p) \in \world{W}$, we set all the values in the $\Th{p}$ sub-sub-block of the $\Th{i}$ sub-block of the $\Th{j}$ block of $\cev{\vctr{a}}_{\vertex{v}_k}$ to $1$. With this assignment, $\sumElemProd{\vec{\vctr{z}}^\timestamp{\:t}_{\vertex{v}_i}, \vec{\vctr{z}}_{\relation{r}_j}, \cev{\vctr{z}}^\timestamp{\,t}_{\vertex{v}_k}}=1$ at $\timestamp{t}=\timestamp{t}_p$. If $(\vertex{v}_i, \relation{r}_j, \vertex{v}_k, \timestamp{t}_p) \in \world{W}^c$, we set all the values for the $\Th{p}$ sub-sub-block of the $\Th{i}$ sub-block of the $\Th{j}$ block of $\cev{\vctr{a}}_{\vertex{v}_k}$ to $-1$. With this assignment, $\sumElemProd{\vec{\vctr{z}}^\timestamp{\:t}_{\vertex{v}_i}, \vec{\vctr{z}}_{\relation{r}_j}, \cev{\vctr{z}}^\timestamp{\,t}_{\vertex{v}_k}}=-1$ at $\timestamp{t}=\timestamp{t}_p$.
\end{proof}

\begin{proposition}
Symmetry, anti-symmetry, and inversion can be incorporated into DE-SimplE in the same way as SimplE.
\end{proposition}
\begin{proof}
Let $\relation{r}_i\in\relations{R}$ with $\function{REMB}(\relation{r}_i)=(\vec{\vctr{z}}_{\relation{r}_i},\cev{\vctr{z}}_{\relation{r}_i})$ be symmetric. According to DE-SimplE, for a fact $(\vertex{v}, \relation{r}_i, \vertex{u}, \timestamp{t})$ we have:
\begin{align}\label{eq:score1}
    \phi(\vertex{v}, \relation{r}_i, \vertex{u}, \timestamp{t}) = \frac{(\sumElemProd{\vec{\vctr{z}}^\timestamp{\:t}_\vertex{v}, \vec{\vctr{z}}_{\relation{r}_i}, \cev{\vctr{z}}^\timestamp{\,t}_\vertex{u}}+\sumElemProd{\vec{\vctr{z}}^\timestamp{\:t}_\vertex{u}, \cev{\vctr{z}}_{\relation{r}_i}, \cev{\vctr{z}}^\timestamp{\,t}_\vertex{v}})}{2}
\end{align}
where $\phi(.)$ gives the DE-SimplE score for a fact, $\vec{\vctr{z}}^\timestamp{\:t}_\vertex{v}$ and $\cev{\vctr{z}}^\timestamp{\,t}_\vertex{v}$ are two vectors assigned to $\vertex{v}$ (according to SimplE) both defined according to Equation~\ref{eq:demb}, and $\vec{\vctr{z}}^\timestamp{\:t}_\vertex{u}$ and $\cev{\vctr{z}}^\timestamp{\,t}_\vertex{u}$ are two vectors assigned to $\vertex{u}$ both defined according to Equation~\ref{eq:demb}. Moreover, for a fact $(\vertex{u}, \relation{r}_i, \vertex{v}, \timestamp{t})$ we have:
\begin{align}\label{eq:score2}
   \phi(\vertex{u}, \relation{r}_i, \vertex{v}, \timestamp{t}) = \frac{(\sumElemProd{\vec{\vctr{z}}^\timestamp{\:t}_\vertex{u}, \vec{\vctr{z}}_{\relation{r}_i}, \cev{\vctr{z}}^\timestamp{\,t}_\vertex{v}}+\sumElemProd{\vec{\vctr{z}}^\timestamp{\:t}_\vertex{v}, \cev{\vctr{z}}_{\relation{r}_i}, \cev{\vctr{z}}^\timestamp{\,t}_\vertex{u}})}{2}
\end{align}
By tying $\vec{\vctr{z}}_{\relation{r}_i}$ to $\cev{\vctr{z}}_{\relation{r}_i}$, the two scores become identical. Therefore, tying $\vec{\vctr{z}}_{\relation{r}_i}$ to $\cev{\vctr{z}}_{\relation{r}_i}$ ensures that the score for $(\vertex{v}, \relation{r}_i, \vertex{u}, \timestamp{t})$ is the same as the score for $(\vertex{u}, \relation{r}_i, \vertex{v}, \timestamp{t})$ thus ensuring the symmetry of $\relation{r}_i$. With the same argument, if $\vec{\vctr{z}}_{\relation{r}_i}$ is tied to $-\cev{\vctr{z}}_{\relation{r}_i}$, then one score becomes the negation of the other score so only one of them can be true.

Assume $\relation{r}_j$ with $\function{REMB}(\relation{r}_j)=(
\vec{\vctr{z}}_{\relation{r}_j},\cev{\vctr{z}}_{\relation{r}_j})$ is known to be the inverse of $\relation{r}_i$. Then the score for a fact $(\vertex{v}, \relation{r}_i, \vertex{u}, \timestamp{t})$ is as in Equation~\eqref{eq:score1} and for $(\vertex{u}, \relation{r}_j, \vertex{v}, \timestamp{t})$ is as follows:
\begin{align}\label{eq:score3}
    \phi(\vertex{u}, \relation{r}_j, \vertex{v}, \timestamp{t}) = \frac{(\sumElemProd{\vec{\vctr{z}}^\timestamp{\:t}_\vertex{u}, \vec{\vctr{z}}_{\relation{r}_j}, \cev{\vctr{z}}^\timestamp{t}_\vertex{v}}+\sumElemProd{\vec{\vctr{z}}^\timestamp{\:t}_\vertex{v}, \cev{\vctr{z}}_{\relation{r}_j}, \cev{\vctr{z}}^\timestamp{\,t}_\vertex{u}})}{2}
\end{align}
By tying $\vec{\vctr{z}}_{\relation{r}_j}$ to $\cev{\vctr{z}}_{\relation{r}_i}$ and $\cev{\vctr{z}}_{\relation{r}_j}$ to $\vec{\vctr{z}}_{\relation{r}_i}$, the score in Equation~\eqref{eq:score3} can be re-written as:
\begin{align}\label{eq:score4}
    \phi(\vertex{u}, \relation{r}_j, \vertex{v}, \timestamp{t}) = \frac{(\sumElemProd{\vec{\vctr{z}}^\timestamp{\:t}_\vertex{u}, \cev{\vctr{z}}_{\relation{r}_i}, \cev{\vctr{z}}^\timestamp{\,t}_\vertex{v}}+\sumElemProd{\vec{\vctr{z}}^\timestamp{\:t}_\vertex{v}, \vec{\vctr{z}}_{\relation{r}_i}, \cev{\vctr{z}}^\timestamp{\,t}_\vertex{u}})}{2}
\end{align}
This score is identical to the score in Equation~\eqref{eq:score1}. Therefore, tying $\vec{\vctr{z}}_{\relation{r}_j}$ to $\cev{\vctr{z}}_{\relation{r}_i}$ and $\cev{\vctr{z}}_{\relation{r}_j}$ to $\vec{\vctr{z}}_{\relation{r}_i}$ ensures $\relation{r}_i$ and $\relation{r}_j$ are the inverse of each other.
\end{proof}

\begin{proposition}
By constraining $\vctr{a}_\vertex{v}$s in Equation~\eqref{eq:demb} to be non-negative for all $\vertex{v}\in\vertices{V}$ and $\sigma$ to be an activation function with a non-negative range (such as ReLU, sigmoid, or squared exponential), entailment can be incorporated into DE-SimplE in the same way as SimplE.
\end{proposition}
\begin{proof}
Let $\relation{r}_i\in\relations{R}$ with $\function{REMB}(\relation{r}_i)=(\vec{\vctr{z}}_{\relation{r}_i},\cev{\vctr{z}}_{\relation{r}_i})$ and $\relation{r}_j\in\relations{R}$ with $\function{REMB}(\relation{r}_j)=(\vec{\vctr{z}}_{\relation{r}_j},\cev{\vctr{z}}_{\relation{r}_j})$ be two distinct relations such that $\relation{r}_i$ entails $\relation{r}_j$. For a fact $(\vertex{v}, \relation{r}_i, \vertex{u}, \timestamp{t})$, the score according to DE-SimplE is as in Equation~\eqref{eq:score1}, and for $(\vertex{v}, \relation{r}_j, \vertex{u}, \timestamp{t})$, the score is as follows:
\begin{align}\label{eq:score5}
    \phi(\vertex{v}, \relation{r}_j, \vertex{u}, \timestamp{t}) = \frac{(\sumElemProd{\vec{\vctr{z}}^\timestamp{\:t}_\vertex{v}, \vec{\vctr{z}}_{\relation{r}_j}, \cev{\vctr{z}}^\timestamp{\,t}_\vertex{u}}+\sumElemProd{\vec{\vctr{z}}^\timestamp{\:t}_\vertex{u}, \cev{\vctr{z}}_{\relation{r}_j}, \cev{\vctr{z}}^\timestamp{\,t}_\vertex{v}})}{2}
\end{align}
By tying $\vec{\vctr{z}}_{\relation{r}_j}$ to $\vec{\vctr{z}}_{\relation{r}_i}+\vec{\vctr{\delta}}_{\relation{r}_j}$ and  $\cev{\vctr{z}}_{\relation{r}_j}$ to $\cev{\vctr{z}}_{\relation{r}_i}+\cev{\vctr{\delta}}_{\relation{r}_j}$, where $\vec{\vctr{\delta}}_{\relation{r}_j}$ and $\cev{\vctr{\delta}}_{\relation{r}_j}$ are vectors with non-negative elements (thus, making this tying scheme equivalent to two inequality constraints), the score in Equation~\eqref{eq:score5} can be re-written as:
\begin{align}\label{eq:score6}
    \phi(\vertex{v}, \relation{r}_j, \vertex{u}, \timestamp{t}) =&~ \frac{(\sumElemProd{\vec{\vctr{z}}^\timestamp{\:t}_\vertex{v}, \vec{\vctr{z}}_{\relation{r}_i} + \vec{\vctr{\delta}}_{\relation{r}_j}, \cev{\vctr{z}}^\timestamp{\,t}_\vertex{u}}+\sumElemProd{\vec{\vctr{z}}^\timestamp{\:t}_\vertex{u}, \cev{\vctr{z}}_{\relation{r}_i} + \cev{\vctr{\delta}}_{\relation{r}_j}, \cev{\vctr{z}}^\timestamp{\,t}_\vertex{v}})}{2} \\
    =&~ \frac{\sumElemProd{\vec{\vctr{z}}^\timestamp{\:t}_\vertex{v}, \vec{\vctr{z}}_{\relation{r}_i}, \cev{\vctr{z}}^\timestamp{t}_\vertex{u}} + \sumElemProd{\vec{\vctr{z}}^\timestamp{\:t}_\vertex{v}, \vec{\vctr{\delta}}_{\relation{r}_j}, \cev{\vctr{z}}^\timestamp{\,t}_\vertex{u}} +
    \sumElemProd{\vec{\vctr{z}}^\timestamp{\:t}_\vertex{u}, \cev{\vctr{z}}_{\relation{r}_i}, \cev{\vctr{z}}^\timestamp{\,t}_\vertex{v}} + 
    \sumElemProd{\vec{\vctr{z}}^\timestamp{\:t}_\vertex{u}, \cev{\vctr{\delta}}_{\relation{r}_j}, \cev{\vctr{z}}^\timestamp{\,t}_\vertex{v}}}{2} \\
    =&~ \phi(\vertex{v}, \relation{r}_i, \vertex{u}, \timestamp{t}) + \frac{\sumElemProd{\vec{\vctr{z}}^\timestamp{\:t}_\vertex{v}, \vec{\vctr{\delta}}_{\relation{r}_j}, \cev{\vctr{z}}^\timestamp{\,t}_\vertex{u}} + \sumElemProd{\vec{\vctr{z}}^\timestamp{\:t}_\vertex{u}, \cev{\vctr{\delta}}_{\relation{r}_j}, \cev{\vctr{z}}^\timestamp{\,t}_\vertex{v}}}{2}
\end{align}
The constraints imposed on the elements of $\vec{\vctr{z}}^\timestamp{\:t}_\vertex{v}, \cev{\vctr{z}}^\timestamp{\,t}_\vertex{v}, \vec{\vctr{z}}^\timestamp{\:t}_\vertex{u},$ and $\cev{\vctr{z}}^\timestamp{\,t}_\vertex{u}$ ensure that all elements of these vectors are non-negative. Furthermore, $\vec{\vctr{\delta}}_{\relation{r}_j}$ and $\cev{\vctr{\delta}}_{\relation{r}_j}$ have also been constrained to be non-negative. 
Therefore, $\sumElemProd{\vec{\vctr{z}}^\timestamp{\:t}_\vertex{v}, \vec{\vctr{\delta}}_{\relation{r}_j}, \cev{\vctr{z}}^\timestamp{\,t}_\vertex{u}}$ and $\sumElemProd{\vec{\vctr{z}}^\timestamp{\:t}_\vertex{u}, \cev{\vctr{\delta}}_{\relation{r}_j}, \cev{\vctr{z}}^\timestamp{\,t}_\vertex{v}}$ are both non-negative resulting in:
\begin{align}\label{eq:score7}
    \phi(\vertex{v}, \relation{r}_j, \vertex{u}, \timestamp{t}) = \phi(\vertex{v}, \relation{r}_i, \vertex{u}, \timestamp{t}) + \frac{\sumElemProd{\vec{\vctr{z}}^\timestamp{\:t}_\vertex{v}, \vec{\vctr{\delta}}_{\relation{r}_j}, \cev{\vctr{z}}^\timestamp{\,t}_\vertex{u}} + \sumElemProd{\vec{\vctr{z}}^\timestamp{\:t}_\vertex{u}, \cev{\vctr{\delta}}_{\relation{r}_j}, \cev{\vctr{z}}^\timestamp{\,t}_\vertex{v}}}{2} \geq \phi(\vertex{v}, \relation{r}_i, \vertex{u}, \timestamp{t})
\end{align}
Since $\phi(\vertex{v}, \relation{r}_j, \vertex{u}, \timestamp{t}) \geq \phi(\vertex{v}, \relation{r}_i, \vertex{u}, \timestamp{t})$, the probability of $(\vertex{v}, \relation{r}_j, \vertex{u}, \timestamp{t})$ being true according to DE-SimplE is greater than or equal to the probability of $(\vertex{v}, \relation{r}_i, \vertex{u}, \timestamp{t})$ being true thus ensuring the entailment of the relations.
\end{proof}

\end{document}